\documentclass[review]{elsarticle}

\usepackage{color}
\usepackage{amsmath,amsfonts,amssymb,pifont}
\usepackage{booktabs}
\usepackage{graphicx}
\usepackage{grffile}
\usepackage{bm}
\usepackage{float}
\usepackage{array}
\usepackage{subfigure}
\usepackage{multirow}
\usepackage{appendix}
\usepackage{lmodern}
\usepackage{makecell}           
\usepackage{multicol}   
\usepackage[colorlinks=true, allcolors=blue]{hyperref}
\usepackage[table,xcdraw]{xcolor}
\usepackage{lineno,hyperref}
\usepackage{wrapfig}

\journal{Journal of \LaTeX\ Templates}

\begin{document}

\begin{frontmatter}

\title{Multi-label Sewer Pipe Defect Recognition with Mask Attention Feature Enhancement and Label Correlation Learning}


\author[a]{Xin~Zuo}
\author[a]{Yu~Sheng}
\author[b]{Jifeng~Shen}
\author[c]{Yongwei~Shan\corref{correspondingauthor}}
\cortext[correspondingauthor]{Corresponding author}
\ead{yongwei.shan@okstate.edu}

\address[a]{School of Computer Science and Engineering, Jiangsu University of Science and Technology, Zhenjiang, 212003, China}
\address[b]{ School of Electronic and Informatics Engineering, Jiangsu University, Zhenjiang, 212013, China}
\address[c]{ School of Civil and Environmental Engineering, Oklahoma State University, Stillwater, OK 74074, USA}


\begin{abstract}
The coexistence of multiple defect categories as well as the substantial class imbalance problem significantly impair the detection of sewer pipeline defects. To solve this problem, a multi-label pipe defect recognition method is proposed based on mask attention guided feature enhancement and label correlation learning. The proposed method can achieve current approximate state-of-the-art classification performance using just 1/16 of the Sewer-ML training dataset and exceeds the current best method by 11.87\% in terms of F2 metric on the full dataset, while also proving the superiority of the model. The major contribution of this study is the development of a more efficient model for identifying and locating multiple defects in sewer pipe images for a more accurate sewer pipeline condition assessment. Moreover, by employing class activation maps, our method can accurately pinpoint multiple defect categories in the image which demonstrates a strong model interpretability. Our code is available at
\href{https://github.com/shengyu27/MA-Q2L}{\textcolor{black}{https://github.com/shengyu27/MA-Q2L.}}
\end{abstract}


\end{frontmatter}


\section{INTRODUCTION}
\label{sec:intro}
The degradation of sewer pipes over time, coupled with various external environmental factors, can lead to structural defects like pipe cracks, collapses, and fractures. These structural failures not only incur substantial economic losses for municipal utilities but also raise significant environmental concerns due to sewage overflows \cite{Bryant2022}. Consequently, timely maintenance and repairs are crucial. Regular inspections for sewer pipe condition assessment are fundamental to proactive asset management strategies. Through these inspections, maintenance and repairs can be scheduled in the early stages of defect occurrence, mitigating the risk of further deterioration.

In the early studies on this topic, sewer pipe inspection heavily relied on Closed-Circuit Television (CCTV) technology to capture video or image data. Subsequently, a trained operator manually annotated multiple defects by viewing the video footage or images. This method is typically inefficient, time-consuming, and labor-intensive \cite{Chen2018}. However, with the development of automation and deep learning techniques in the field of computer vision, it is now feasible to utilize sensors or cameras to capture pictures or videos of sewer pipes. Deep learning algorithms are then employed to detect defects, and the detection results are presented to trained personnel for feedback \cite{Haurum2020}, \cite{Li2022}.

Given the inherent characteristics that multiple manually labeled defects may coexist in sewer pipe images \cite{Haurum2020}, multi-label classification techniques, capable of recognizing multiple defects simultaneously, are well-suited for such tasks. Moreover, in recent years, image multi-label classification techniques have garnered considerable success in the realm of deep learning. The method by Tao et al. \cite{Tao2022} employs channel attention and spatial attention to gather the global contextual information of images and improve label correlation by constructing a model utilizing a graph attention network. However, this approach necessitates a two-stage procedure, which is time-consuming and difficult. Furthermore, channel attention may result in the loss of fine-grained position information, resulting in a failure to sufficiently collect position-related feature information. Another study \cite{Haurum2022} presents a multi-scale fusion-based end-to-end multi-scale hybrid vision transformer for modeling non-local spatial semantics. However, this multi-scale fusion is not suitable for images with multiple classes of defects coexisting, because pooling and down-sampling in the fusion operation result in the loss of the defect details and local information, rendering the model incapable of learning that information. To solve this issue, a mask attention method is proposed that focuses on the defective area information in order to increase the discriminative capacity of local spatial feature information and the classification performance of the model.

Due to the intricate nature of the sewer pipe environment, special attention needs to be directed towards the unique characteristics of the data resource. Sewer pipe defect images are derived from real-time video and are subject to post-processing. The defect categories in the images are diverse, with random features that are impossible to forecast manually. As a result, ensuring an equal distribution of defect categories in the final dataset is unachievable, resulting in a severe class imbalance issue known as the long-tail problem \cite{Zhang2023}. It is also illustrated in the literature \cite{Haurum2021} that there is indeed a long-tail problem with sewer pipe defect images, and it is also a continuing concern in the field of multi-label classification. To overcome this issue, an asymmetric loss function is used, and targeted category weight values are built to allow the model to alter parameters during backpropagation to improve learning ability for categories with fewer samples and more difficult classification tasks.

Furthermore, due to the variety of defects, one pipe image may be showing numerous problems at the same time. The coexistence phenomenon \cite{Pu2023} can be clearly observed by analyzing Fig. 2(a) and calculating the occurrences of a specific defect category in conjunction with other specified defect categories in the Sewer-ML dataset. For example, when a crack defect occurs, it is common for a joint error or surface damage defect to also exist. Fig. 2(b) shows examples of images of defective sewer pipe. The upper image shows the following defects at the same time: cracks, breaks, and collapses (RB); surface damage (OB); displaced joint (FS); and connection with constructure changes (OK). Note that the 'Cracks' in Fig. 2(a) refer to RB defects. The lower picture shows defects such as cracks, breaks, and collapses (RB), displaced joint (FS), obstacle (FO), and branch pipe (GR). According to a qualitative study, pipe cracks are often produced by external forces or protracted corrosion owing to a wet environment and fluid transit \cite{Bryant2022}. As a result, the inner wall of a pipeline is frequently damaged, resulting in surface damage problems. Serious crack defects, if not maintained or repaired in a timely manner, can cause pipe fracture or even collapse, with the collapsed pipe material becoming an obstacle inside the pipe. To address the issue, self-attention computation is employed to help the model pay greater attention to the relevant information between labels to enhance label relevance and improve model performance.

\begin{figure}[h]
	\centering
	\includegraphics[width=\textwidth]{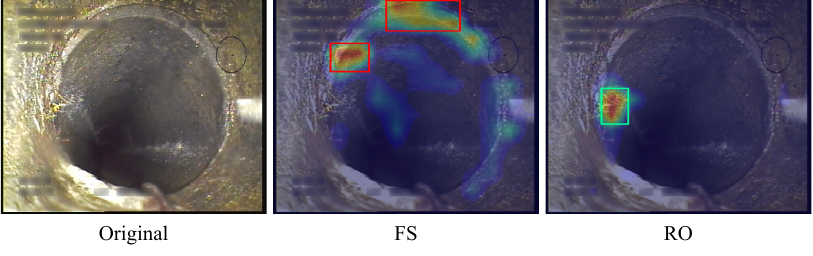}
	\caption{The method in this paper can roughly localize the defect information in the image. The given example contains both types of defects: FS: displaced joint; RO: roots;}
	\label{Fig.1}
\end{figure}

\begin{figure}[h]
	\centering
	\subfigure[]{
		\includegraphics[width=7.08cm,height=6.62cm]{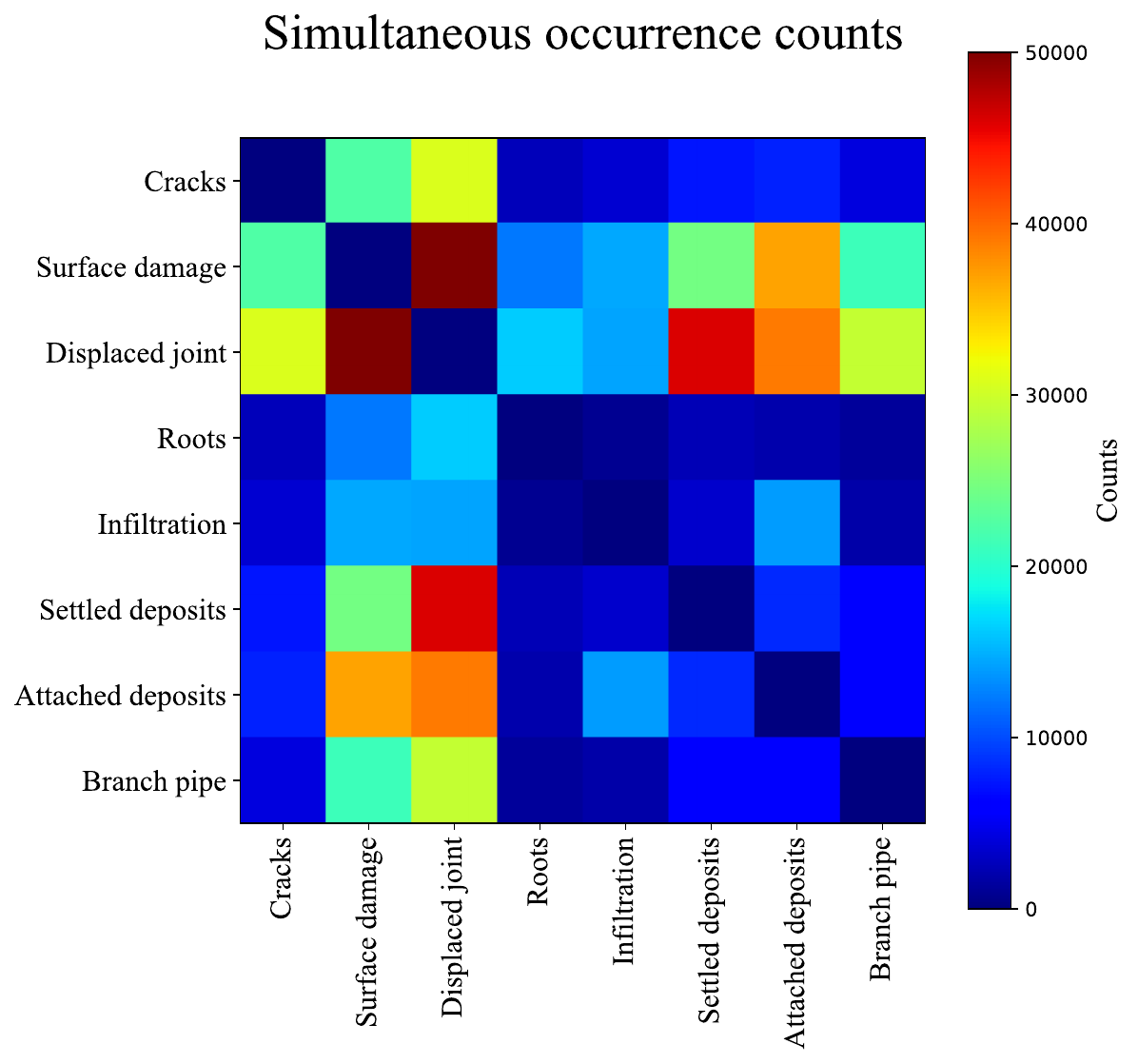} 
	}
	\quad
	\subfigure[]{
		\includegraphics[height=6.62cm]{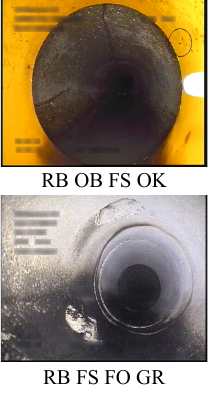}  
	}
	\caption{(a) Coexistence of multiple classes of defects obtained from the Sewer-ML dataset. (b) Some examples.}
	\label{Fig.2}
\end{figure}

To address these issues mentioned above, the authors propose a Mask Attention guided Q2L method (MA-Q2L) for multi-label sewer pipe defect recognition in complex sewer pipe environment. Firstly, a mask attention guided feature enhancement module is introduced, which leverages class activation map to obtain the local spatial discriminative information of different defects. The utilization of class activation map is preferred because they generate spatial discriminative information that is visualizable, providing further evidence that using class activation maps to explore local spatial feature information is effective. Secondly, the self-attention mechanism is incorporated to model the relationship between the labels, thereby improving label relevance. Since interaction learning between label embeddings and image characteristics may fail to capture the relationship and information between the labels, the self-attention computation allows the model to focus more on the correlated information between the labels, making the relationship between various labels clearer and more distinct. Finally, a specialized method based on the asymmetric loss function is given, which designs targeted category weight values. This allows the model to adjust parameters during the back propagation process, enhancing the learning capacity for categories with fewer samples and more challenging classification. The traditional asymmetric loss function struggles to effectively deal with the long-tail problem arising from a severe category imbalance in sewer pipe defect images while dynamically acquiring weight values to address potential inflexibility in the weight value acquisition process. Our proposed method has achieved the state-of-the-art performance on only 1/16 of the sewage pipe dataset Sewer-ML, and even 73.15\% on the entire dataset, representing a significant improvement in model performance over prior models. In addition, as shown in Fig. 1, the method in this paper can localize the defect information in the image more accurately. 

The primary contributions of this paper are as follows:

\begin{itemize}
	\item An attention mask guided spatial feature enhancing method is proposed to improve the local feature discriminative ability.
	\item The relevance of labels is improved through the self-attention mechanism of label embedding.
	\item An asymmetric loss function with static and dynamic weight update technique is developed to relieve the class imbalance issue.
	\item Our model not only achieves state-of-the-art (SOTA) performance on the multi-label classification task, but is also able to roughly localize pipe defects in the image, which demonstrates strong model interpretability.
	
\end{itemize}

\section{Related work}
\subsection{Automated sewer inspections}
For decades, researchers and industry professionals have dedicated efforts to automatic sewer pipe defect recognition, aiming to achieve breakthroughs and develop efficient models for real-world applications. However, according to Haruum and Moeslund \cite{Haurum2020}, progress in this field has been slow, attributed to limitations in datasets and a scarcity of freely available modeling code. For the automated recognition and categorization of sewage pipe defects, previous studies predominantly utilized Closed-Circuit Television (CCTV) images \cite{Chen2018}, \cite{Haurum2021}, \cite{PuTao2023} and other sensor-based approaches \cite{Alejo2017}, \cite{Bahnsen2021}, \cite{haurum2021sewer}. Furthermore, the lack of standardized datasets and assessment measures \cite{Xie2019}, \cite{Chen2022}, \cite{Dang2022}, has impeded meaningful comparisons between various models, making the evaluation of their generalization effectiveness challenging \cite{Haurum2020}. 

After the release of the Sewer-ML dataset \cite{Haurum2021}, several researchers, such as Tao \cite{Tao2022} and Haurum \cite{Haurum2021}, devised two-stage classification models. However, due to the intricacies and suboptimal performance of these two-stage models, this approach gradually lost favor. Haurum et al. \cite{Haurum2022} subsequently introduced an end-to-end network model, showcasing commendable performance. Nevertheless, this technique lacked in-depth exploration and utilization of the local spatial feature information within the original image. Concurrently, alternative methods \cite{haurum2022multi} initiated multi-task classification, utilizing a graph neural network decoder to refine the prediction of each classification and identify issues and attributes related to sewer pipes. However, in terms of the model structure design, the defect classification received support primarily from information related to the other tasks, neglecting the discriminative local spatial feature within the image. Considering the specific characteristics of sewer pipes, additional approaches \cite{Hu2023} aimed to diminish defect ambiguity by refining sewer pipe defects in the latent space, ultimately improving the performance of defect recognition and classification. 

While these technologies introduced novel perspectives, they often overlooked the inherent local spatial features of the original image. Our method extracts this information by generating the attention mask using the class activation map, which is then incorporated into the attention computation to enhance the learning capabilities of the model. This strategy facilitates a thorough exploration and utilization of the native local spatial feature present in the original image.

In addition to image-level defect recognition methods, there are also some pixel-level segmentation-based approaches. Mask R-CNN\cite{he2018mask}, for instance, incorporates an additional branch network, employing ROIAlign to isolate segmentation targets and predict binary masks for each category independently. Sophisticated models, such as large model Segment Anything\cite{2023segment}, achieve remarkable segmentation performance on extensive datasets through prompt engineering. These models leverage a large number of manually annotated segmentation masks and achieve high-performance segmentation through loss constraints within the model. However, the Sewer-ML dataset utilized in this study lacks pixel-level segmentation masks. Therefore, the method proposed in this article relies on the model's efficient feature extraction capabilities to achieve effective results, producing rough segmentation renderings without relying on segmentation masks.

\subsection{Multi-label classification}
Recent studies in the domain of multi-label classification have primarily concentrated on two key aspects to improve and optimize model performance. On one hand, efforts were made to optimize model performance using spatial features. Object detection methods, for instance, were initially utilized to pinpoint features, followed by subsequent feature extraction \cite{Nie2022}, \cite{Wei2015}. While this approach alleviated the complexity of feature extraction, acquiring prior information became more challenging, necessitating manual filtering and labeling of the labels, consuming substantial resources \cite{Demidov2023}. This proved unnecessary and undesirable for a classification model. Other researchers have explored the correlation between spatial features and labels \cite{Pu2023}, \cite{Xu2022}, \cite{Zhu2022}, linking them through model structure design. However, this approach required a substantial volume of labeled data to support model training, demanding an intricately designed model structure. In scenarios with limited data, the performance of the model would experience a significant downturn. Additionally, certain algorithms \cite{Lanchantin2021}, \cite{Zhou2023} aimed to classify images utilizing spatial location occlusion masks or multi-scale feature fusion. While the model could grasp finer-grained features using diverse location-based occlusion masks, in the presence of a more concentrated feature set, a considerable portion of features would be masked, leading to substantial loss of feature information. Moreover, in the process of multi-scale feature fusion, pooling and up-sampling contributed to the loss of information regarding local features.

On the other hand, researchers started delving into the correlation between labels. In multi-label classification dataset images, there might be a co-occurrence between labels. To address this phenomenon, linking these labels through graph neural networks or label embedding to enhance the correlation between labels \cite{Liu2021}, \cite{Wu2022}, \cite{Yuan2023} could enhance the classification accuracy of the model. However, this required considering the intricate connections between the labels of the dataset. For complex and specialized datasets, a more sophisticated design might be necessary, introducing more burdensome weights and computations.

Similarly, the authors embarked on a search in two aspects. Firstly, a class activation map is utilized to generate the attention mask, enhancing the capacity of the model to learn diverse spatial features related to defects. This is akin to obtaining prior information from the class activation map, which is substantially less expensive than using object detection method. Secondly, for label relevance, label embedding is employed for self-attention computation, enhancing the learning capacity of the model for label-to-label interactions.

\subsection{Loss function and long-tail problem}
The cross-entropy loss function \cite{chen2019learning}, \cite{chen2019multi} was commonly utilized in multi-label classification, contributing to a significant performance gap across labels due to the inherent long-tail issue in multi-label datasets. Subsequently, focal loss \cite{lin2017focal} was inextricably linked to multi-label categorization. Focal loss augmented cross-entropy with configurable parameters to alter the balance between positive and negative samples, allowing the model to focus more on challenging-to-classify negative samples for dataset balance and enhanced performance. However, it controlled both positive and negative samples with the same parameter, increasing the contribution of negative samples while decreasing the contribution of positive ones.

The ASL(Asymmetric Loss) \cite{Ridnik2021} decoupled the parameters of focal loss, adjusting the learning capacity of the model for positive and negative samples separately, rendering it more appropriate for datasets with the long-tail issue. Some researchers have already employed ASL for multi-label classification \cite{Xu2022}, \cite{Zhu2022}, \cite{Liu2021} with promising results. However, the ASL parameters were statically specified and might not be adequately fitted to a specific dataset. 

Considering the presence of the long-tail problem in sewer pipe defect data, the authors employ ASL. Simultaneously, the weight values are elevated for bottleneck categories characterized by fewer samples and greater classification difficulty. During backpropagation, the model parameters are adjusted to facilitate enhanced learning for these bottleneck categories. This method can better deal with the problem while retaining the original control for the contribution of positive and negative samples, thereby improving the overall performance of the model. This approach straightforwardly and effectively addresses the issue while preserving the original control over the contribution of positive and negative samples, leading to an overall improvement in model performance.

\section{The Proposed Method}
\begin{figure}[h]
	\centering
	\includegraphics[width=\textwidth]{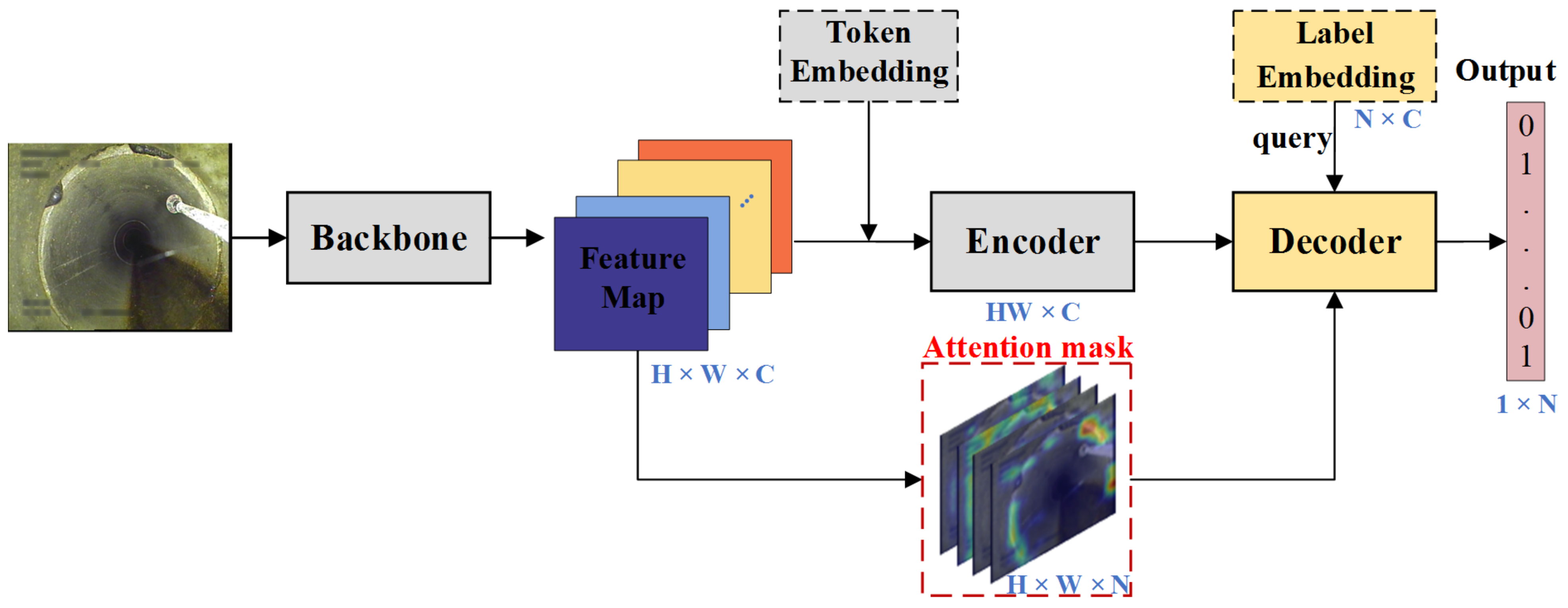}
	\caption{The overall structure of the proposed method. Where H, W are the height and width of the image after feature extraction, C is the number of channels and N is the number of dataset categories. In our model H=14, W=14, C=2048, N=17.}
	\label{Fig.3}
\end{figure} 

In this section, firstly, the overall structure of a model is given (Fig. 3), which is inspired by Q2L \cite{Liu2021}. Meanwhile, the authors chose Q2L as our baseline since it has commendable performance on public multi-label classification datasets. Our model uses Resnet101 as the backbone and then employs the classic Transformer structure, the encoder-decoder structure, to combine label embedding for interactive learning. An attention mask, derived from the backbone using a class activation map, is then integrated into the decoder. The label embedding information undergoes linear mapping to achieve multi-label image classification. In contrast to the Q2L baseline, our model substitutes the original backbone with an improved backbone, integrates a custom-designed attention mask into the encoder, and adopts an enhanced asymmetric loss function in lieu of the original loss function. Subsequent sections will delve into the detailed explanation of each module.

\subsection{Problem formulation}

Given an input image {$\bm{I} \in \mathbb{R}^{3 \times h \times w}$}, the authors aim to design a model for predicting a multi-label vector $ \bm{O} \in \left\{ 0,1 \right\}^{1 \times N}$ representing the defect categories present in the image. In this binary encoding, 1 indicates the presence of the corresponding defect category, while 0 signifies its absence. The model comprises three functions, $F_{backbone}$, $F_{encoder}$ and $F_{decoder}$. The function $F_{backbone}$ is employed to extract features from the raw input image to obtain the feature representation $\bm{X} \in \mathbb{R}^{C \times H \times W}$. The function $F_{encoder}$ processes these features further and captures the information between image features to get the encoded feature $ \bm{X_{e}} \in \mathbb{R}^{C \times HW}$.  $F_{decoder}$ interacts with the label embedding which directly passes the label through the embedding operation and extracts its weight value, denoted as $\bm{LE}$, to learn from the extracted features $\bm{X_{e}}$ and ultimately produces the predicted result $\bm{O} \in \mathbb{R}^{1 \times N}$. The specific definitions of these functions can be denoted as:

\begin{gather}
	\label{eq1}
	\bm{X} = F_{backbone}(\bm{I}) \\
	\label{eq2}
	\bm{X_{e}} = F_{encoder}(\bm{X}) \\ 
	\label{eq3}
	\bm{O} = F_{decoder}\left( \bm{X_{e}},\bm{LE} \right) 
\end{gather}
where $h$ and $w$ are the height and width of the input image $\bm{I}$, respectively, $H$, $W$ and $C$ are the height, width, and channel number of the feature map. $N$ is the total number of defect categories.

\subsection{Encoder}
\label{section:3.2}
The authors adopt a convolutional structure as the backbone, which can model and learn local information. In order to capture the global context semantic feature information, the features are fed into the encoder structure and further processed to get high-level semantic features. The encoder includes a multi-head self-attention mechanism, enabling the learning of multiple distinct feature representations. Each attention head focuses on different local information within the input sequence or features, capturing richer semantic information. Specifically, $h$ denotes the number of heads, and different projection matrices are used to generate different tensors: \bm{$\left. \left\{ Q \right._{i} \right\}_{i = 1}^{h}$}, \bm{$\left. \left\{ K \right._{i} \right\}_{i = 1}^{h}$}, \bm{$\left. \left\{ V \right._{i} \right\}_{i = 1}^{h}$}. Multi-head self-attention can be formulated in Eq. (4)-(6).
\begin{gather}
	\label{eq4}
	Attention\left( {\bm{Q},\bm{K},\bm{V}} \right) = softmax\left( \bm{{QK^{T}} / \sqrt{\bm{d_{k}}}} \right)\bm{V} \\
	\label{eq5}
	MultiHead\left( {\bm{Q^{'}},\bm{K^{'}},\bm{V^{'}}} \right) = Concat\left( \bm{{head}_{1}},\ldots,\bm{{head}_{h}} \right)\bm{W} \\ 
	\label{eq6}
	{head}_{i} = Attention\left( \bm{{Q_{i}},\bm{K_{i}},\bm{V_{i}}} \right)
\end{gather}
where $\bm{Q^{'}}$ is the cascade of $\bm{\left. \left\{ Q \right._{i} \right\}_{i = 1}^{h}}$, \bm{$K^{'}$} and \bm{$V^{'}$} are also the same settings, and \bm{$W$} is the linear projection matrix. The whole encoder structure can be represented as:
\begin{gather}
	\label{eq7}
	\bm{F_{0}} = MultiHead\left( TE(\bm{X}),TE(\bm{X}),TE(\bm{X}) \right) \\
	\label{eq8}
	\bm{F_{1}} = Norm\left( \bm{X} + dropout\left( \bm{F_{0}} \right) \right) \\ 
	\label{eq9}
	\bm{F_{enc}} = {Norm\left( \bm{F \right.}_{1}} + dropout\left( FFN\left( \bm{F_{1}} \right) \right))
\end{gather}
where $TE$ stands for token embedding, $Norm$ and $dropout$ stand for normalization operations to prevent overfitting, $FFN$ stands for feed-forward network, and \bm{$F_{enc}$} stands for the final feature representation of the encoder structure, where the positional information needs to be added in the attention computation, which is not represented in Eq. 7 for clarity.

\subsection{Attention mask}

In addition to entering the encoder structure, another path for features from the backbone is to generate the attention mask through class activation map. Let’s first introduce the process of generating the class activation map: the feature map $\bm{X} \in \mathbb{R}^{C \times H \times W}$ is obtained from backbone. Then, after global average pooling, the height and width dimensions of the feature map are averaged to yield a global feature vector. Next, the vector is linearly mapped to the same dimensions as the number of defect categories to obtain the weights which represents local spatial feature information about the defects in the corresponding image. Subsequently, the weights are fused with the original feature map to generate the class activation map, serving as the attention mask. 
Specifically, for each pixel in the image, the model assigns different values according to different defect categories, indicating whether the model pays attention to the pixel or not. Larger values imply greater model attention, suggesting a higher likelihood of the presence of the defect category at that location. Conversely, smaller values mean that the model pays less attention to the pixel, considering it less relevant information. The process of generating the weights is depicted in Eq. 10:
\begin{gather}
	\label{eq10}
	\bm{A_{0}} = ~L_{\bm{W}}\left( GAP(\bm{X}) \right)
\end{gather}
where \bm{$X$} represents the feature map obtained from backbone, $GAP$ is global average pooling, $L$ is linear mapping, and \bm{$W$} represents the weights of the linear mapping layer. \bm{$A_{0}$} is the classification probability result, which is not needed. The authors only need the weights $W$ which are fused with the feature map \bm{$X$} to generate the attention mask, as shown in Eq. 11:
\begin{gather}
	\label{eq11}	
	\bm{A_{1}} = \bm{~X}\ast \bm{W}
\end{gather}
where $\bm{W} \in \mathbb{R}^{C \times N}$, $\bm{A_{1}} \in \mathbb{R}^{HW \times N}$ is the attention mask.
The attention mask is integrated into the subsequent computation of the attention mechanism, thereby augmenting the learning capability of the model. Importantly, this process is akin to acquiring prior information from the class activation map, contributing to enhanced performance in the defect recognition classification task.

\subsection{Decoder}

To capture information from the labels, the decoder structure is followed by the encoder, enabling interactive learning between the label embeddings and the features from the encoder to enhance model performance. The label embedding setting of the model is adopted from Q2L to obtain the query and its iterative update method. In the cross-attention of decoder, the key and value come from the encoder \bm{$F_{enc}$}, while the query originates from the label embedding \bm{$LE$}, representing the label information. Cross-attention enables dynamic interaction between label embeddings and image features, effectively modeling the relationship between features and labels. Ultimately, the query learns the presence of specific categories in the image, facilitating binary classification decisions.

At the same time, the attention mask is integrated into cross-attention. The information in the attention mask is added to the results of query and key computations, prompting the model to pay more attention to the meaningful defect local spatial discriminative information region during interactive learning. This approach enhances the accuracy of defect detection in the image, contributing to overall model performance improvement. Moreover, relying solely on the interactive learning between image features and label embeddings may not capture the intricate label relationships adequately. To address this, self-attention computation is employed, making connections between different labels more explicit and distinct. Therefore, this enhancement learning for label relevance by performing self-attention computation is shown in Fig. 4, which is structured as:

\begin{gather}
	\label{eq12}	
	{CA}_{i}\left(\bm{ Q_{i}},\bm{K},\bm{V},\bm{A_{1}} \right) = Softmax\left( \bm{{Q_{i}K^{T}}/\sqrt{\bm{d_{k}}} + \bm{A_{1}}} \right)\bm{V} \\
	\label{eq13}
	\bm{Q_{t}} = Norm\left( {\bm{Q_{i}}{ + dropout\left( CA \right.}_{i}\left( \bm{Q_{i}},\bm{K},\bm{V},\bm{A_{1}} \right))} \right) \\
	\label{eq14}
	\left. \bm{Q_{t + 1}} = {Norm\left( \bm{Q_{t}} + dropout\left( MultiHead\left( \bm{Q \right. \right. \right.}_{t}},\bm{Q_{t}},\bm{Q_{t}} \right))) \\
	\label{eq15}
	\left. \bm{Q_{i + 1}} = {Norm\left( \bm{Q_{t + 1}} + dropout\left( FFN\left( \bm{Q \right. \right. \right.}_{t + 1}} \right)))
\end{gather}
where, \bm{$Q_{i}$} represents the query in the i-th decoder. The initial query, \bm{$Q_{1}$}, is obtained from label embedding \bm{$LE$}. \bm{$K$} represents the key, and \bm{$V$} represents the value, both come from the encoder \bm{$F_{enc}$}. \bm{$A_{1}$} represents the attention mask. ${CA}_{i}$ represents the output of the i-th cross-attention calculation. \bm{$Q_{i + 1}$} represents the output of the whole decoder, which also serves as the input of the next query of decoder. \bm{$Q_{t}$} and \bm{$Q_{t+1}$} represent the temporary variables required in the decoder after self-attention computation in Eq. (13)-(15). Among them, the attention computation needs to incorporate the position information, which is not represented in Eq.14.

\begin{figure}[h]
	\centering
	\includegraphics[scale=0.27]{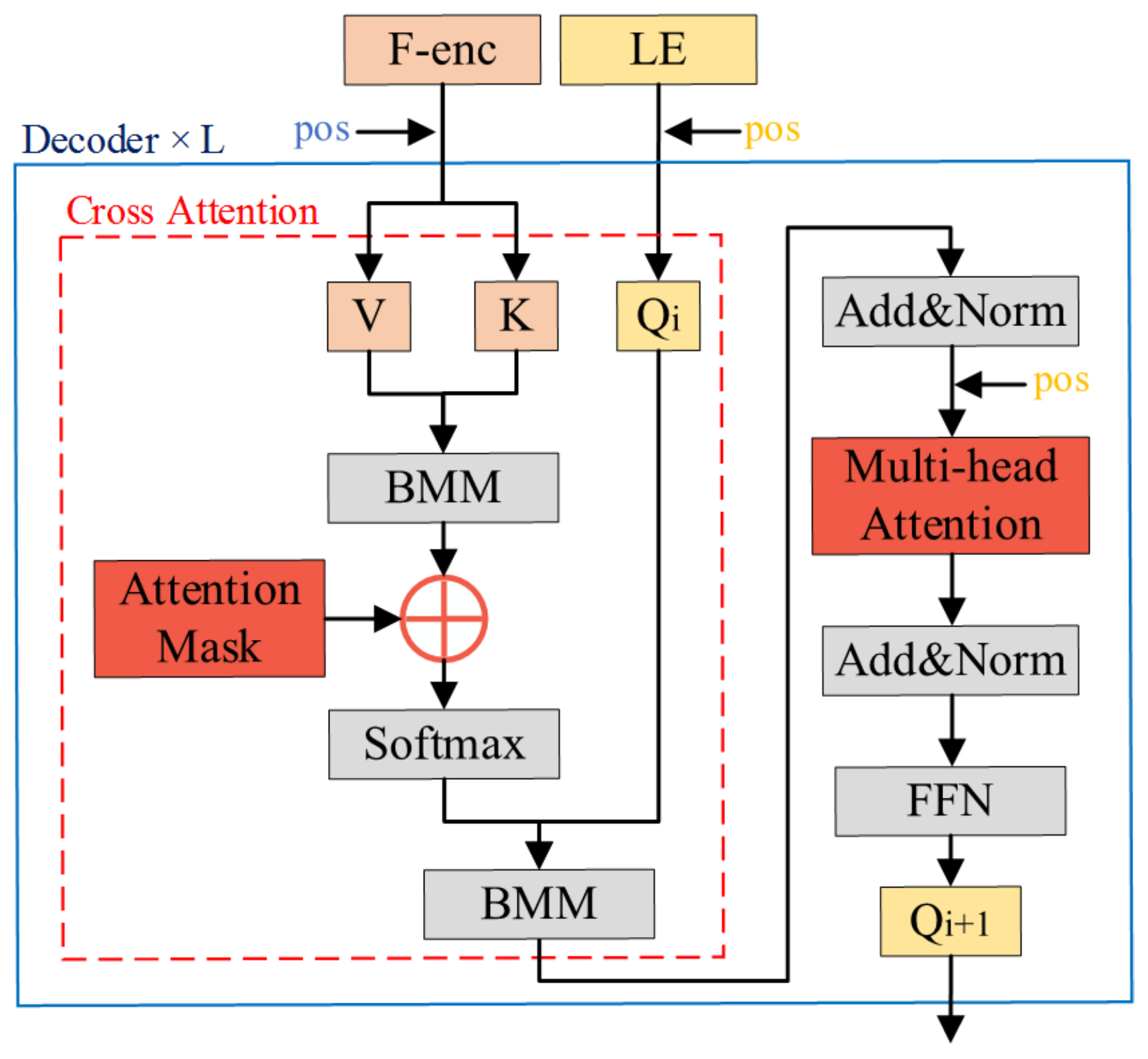}
	\caption{The architecture of the decoder.}
	\label{Fig.4}
\end{figure}

\subsection{Static loss weight}
Our model employs a targeted category weighting technique based on an asymmetric loss function to effectively deal with the long-tail problem in sewer pipe defect data. In multi-label classification, it is customary to simplify the problem into a series of binary classification tasks. Given $N$ labels, the basic network produces one logit value per label, \bm{$Z_{n}$}. Each logit is independently activated by a sigmoid, and \bm{$y_{n}$} serviced as the ground truth for the category $n$. The total classification loss, $L_{total}$, can be obtained by aggregating the binary classification losses from the $N$ labels. As shown in Eq.16.
\begin{gather}
	\label{eq16}	
	L_{total} = {\sum\limits_{n = 1}^{N}L}\left( sigmoid\left( \bm{Z_{n}} \right),\bm{y_{n}} \right)
\end{gather}
The general form of the bisection loss $L$ for each label is shown in Eq.17:
\begin{gather}
	\label{eq17}	
	L = - \bm{yL_{+}} - (1 - \bm{y})\bm{L_{-}}\\
	\label{eq18}	
	\left\{ {~\begin{matrix}
			{\bm{L_{+}} = (1 - \bm{p})^{\gamma}{\log(\bm{p})}} \\
			{\bm{L_{-}} = \bm{p}^{\gamma}{\log(1 - \bm{p})}}
	\end{matrix}} \right.
\end{gather}
where \bm{$y$} is the ground truth label (category index $n$ is omitted for convenience), and \bm{$L_{+}$} and \bm{$L_{-}$} denote the positive and negative loss components, respectively. $\gamma$ is a parameter that controls the extent of learning of the model for positive and negative samples, and $\bm{p} = sigmoid\left( \bm{Z_{n}} \right)$. Focal loss is introduced to address the long-tail problem of the dataset, and controlling $\gamma$ allows the model to focus on learning the categories that are challenging to categorize, thereby amplifying the contribution of negative samples and diminishing the contribution of positive samples.

ASL decouples $\gamma$ in the original focal loss into two parameters, $\gamma_{+}$ and $\gamma_{-}$, as a means of controlling the learning degree of the model for positive and negative samples, respectively. ASL reduces the contribution of negative samples to the loss by thresholding m, solving the problem of insufficient contribution to positive samples presented in the focal loss. Experimental evidence \cite{Ridnik2021} has demonstrated that setting  $\gamma_{+} = 0$ and $\gamma_{-}$ to a static value of 2 yields the best result compared to the dynamic value.

On top of ASL, the authors introduce multiplicative weights $\alpha$ to the binary loss function $L$ for each label. This adjustment increases the loss share of the bottleneck categories when calculating the overall loss, channeling the learning of the model focus towards these specific categories. This concept parallels the design of ASL for $\gamma_{-}$, which tunes the contribution of negative samples to the overall loss, accentuating challenging negative samples. Similarly, our $\alpha$ aims to modulate the learning capacity of the model for the bottleneck categories, regulating their contribution to the overall loss. For easily recognizable categories, $\alpha$ remains at its original weight of 1, while for harder-to-recognize categories, $\alpha$ is set to 2.

By introducing the weight updating strategy of $\alpha$, the authors enhance the learning for the bottleneck categories. This approach addresses the loss between different categories in a more balanced manner, improving the ability of the model to recognize bottleneck categories and further enhancing the performance of the sewer pipe defects classification task. The specifics are as follows:
\begin{gather}
	\label{eq19}	
	L_{total} = {\sum\limits_{n = 1}^{N}{\alpha L}}\left( {sigmoid\left( \bm{Z_{n}} \right),\bm{y_{n}}} \right)\\
	\label{eq20}	
	\left\{ {~\begin{matrix}
			{\alpha = 1,~~n \notin B} \\
			{\alpha = 2,~~n \in B}
	\end{matrix}} \right. \\
	\label{eq21}
	B = \left\{ {RB,IS,FO,OS} \right\}
\end{gather}
where $n$ represents a certain category and $B$ represents the set of bottleneck categories, which contains the four defect categories, and the selection of this set is described in experimental section 4.3.

\subsection{Dynamic loss weight}
In refining our loss function, the hyperparameters are typically tuned through manual experimentations, which is a laborious and time-consuming process. Specifically, determining the suitable $\alpha$ weight poses a challenge. To address this, a dynamic and learnable method for the values of $\alpha$ is introduced. The authors gauge the difficulty of recognizing and classifying each category by computing its mean average precision ($mAP$). This assessment relies on the variance between the $mAP$ of each category and the overall mean $mAP$, scaled to a single-digit value, which is then employed as the $\alpha$ value. Specifically:
\begin{gather}
	\label{eq22}	
	{Mean}_{mAP} = \frac{1}{N}{\sum\limits_{n = 1}^{N}{mAP}_{n}}\\
	\label{eq23}	
	\mathrm{\Delta}{mAP}_{n} = {Mean}_{mAP} - {mAP}_{n} \\
	\label{eq24}	
	\alpha_{n} = \mathrm{\Delta}{mAP}_{n}/10
\end{gather}
where ${mAP}_{n}$ represents the $mAP$ value of $n$ category. ${Mean}_{mAP}$ represents the mean value of all categories. $\mathrm{\Delta}{mAP}_{n}$ represents the difference between the $mAP$ of $n$ category and the mean value of the $mAP$ of all categories, and $a_{n}$ represents the weight value to be added to $n$ category. The algorithm, using difference calculation, can be seen as evaluating the gap between each category and the mean value, reflecting the overall performance of each category under the training of the model. After this, the authors sort $a_{n}$ in descending order, taking the first $m$ corresponding weight values set to the weighted values of the corresponding categories, and setting the other values to 1, allowing the model to weight only the first $m$ categories that were not learned well enough, while keeping the focus on the other categories unchanged, specifically:
\begin{gather}
	\label{eq25}	
	\alpha_{n} = \left\{ \begin{matrix}
		{\alpha_{n},~n \leq m} \\
		{1,n > m}
	\end{matrix} \right.
\end{gather}
where $m$ represents the number of top values to be selected for category weights after descending sorting of weights. 

Considering that the evaluation metrics are ${F1}_{Normal}$ and ${F2}_{CIW}$, experimental analysis revels that the sequence and magnitude of $mAP$ values for each category align with those of ${F1}_{Normal}$ and ${F2}_{CIW}$. Hence, opting for the mean of $mAP$ values is consistent with choosing the mean of either ${F1}_{Normal}$ or ${F2}_{CIW}$ values. The selection of either method yields an equivalent effect.

\section{Experiment}
In the experimental section, the authors commence by detailing the experimental setup, followed by outlining the process for selecting the one-sixteenth sub-dataset of Sewer-ML. Subsequently, the authors elucidate the experiments conducted to identify the bottleneck categories, concluding with the presentation of results where our model is compared to the baseline as well as to other models.

\subsection{Experiment setup}
\subsubsection{Experiment details}
In constructing our model, the authors adhere to the settings in the Q2L model. The backbone of our model is Resnet101, pre-trained on ImageNet, while the encoder and decoder utilize the official Transformer structure modules \cite{Vaswani2017}. All images are resized to 448 $\times$ 448 as the input to the model. The feature dimension after backbone extraction is H $\times$ W $\times$ C = 14 $\times$ 14 $\times$ 2048. For training, the authors employ the Adam optimizer \cite{Kingma2014} with a learning rate of 1e-5, a weight decay of 1e-2, a batch size of 16, and 40 epochs of iterative training. Additionally, the authors apply an exponential moving average (EMA) to model parameters with a decay of 0.9997. Our code will be made publicly available, and more detailed settings can be found in the code repository.

\subsubsection{Evaluation metrics}

Following the evaluation metrics in \cite{Haurum2021}, ${F1}_{Normal}$ and ${F2}_{CIW}$ are used as metrics to evaluate the model performance. In order to bring in domain knowledge, they both follow $F_{\beta}$:

\begin{gather}
	\label{eq26}	
	F_{\beta} = (1 + \beta^{2})Precision \times Recall/\left( \beta^{2} \times Precision + Recall \right) \\
	\label{eq27}
	Precision = ~TP/(TP + FP) \\
	\label{eq28}
	Recall = ~TP/(TP + FN)
\end{gather}

where, $\beta$ is the weight of recall and $F_{\beta}$ considers recall to be $\beta$ times more important than accuracy. $TP$ and $FN$ represent the number of correctly or incorrectly detected defective samples, while $FP$ denotes the number of non-defective samples incorrectly classified as defective. In the sewer pipe defect classification task, the metric of normal image is denoted as ${F1}_{Normal}$, i.e., $\beta = 1$. For defective images, the class importance weight (CIW) needs to be incorporated into the calculation, as illustrated in Table 1. Moreover, because $FN$ has a greater economic impact than $FP$ \cite{Pu2023}, $\beta = 2$ needs to be set.
\begin{gather}
	\label{eq29}
	{F2}_{CIW} = \left( {\sum\limits_{n = 1}^{N}{{F2}_{n} \times {CIW}_{n}}} \right) / {\sum\limits_{n = 1}^{N}{CIW}_{n}}
\end{gather}
where ${F2}_{n}$ and ${CIW}_{n}$ are the $F2$ and $CIW$ values of $n$ categories, respectively, and $N$ is the total number of categories.

\begin{table}
	\centering
	\caption{Overview and short description of each annotation class \protect\cite{og2010fotomanualen} and the class-importance weights(CIW) \protect\cite{ogfotomanualen}. Reproduced from Haurum and Moeslund \protect\cite{Haurum2021}.}
	\label{Table1}
	\begin{tabular}{clc}
		\toprule
		Code & Description & CIW \\
		\midrule
		RB & Cracks, breaks, and collapses & 1.0000  \\
		OS & Lateral reinstatement cuts & 0.9009 \\
		FS & Displaced joint & 0.6419 \\
		OB & Surface damage & 0.5518  \\
		OK & Connection with construction changes & 0.4396 \\
		PH & Chiseled connection & 0.4167 \\
		PB & Drilled connection & 0.4167 \\
		OP & Connection with transition profile & 0.3829 \\
		RO & Roots & 0.3559 \\
		IN & Infiltration & 0.3131 \\
		PF & Production error & 0.2896  \\
		FO & Obstacle & 0.2477 \\
		BE & Attached deposits & 0.2275 \\
		IS & Intruding sealing material & 0.1847 \\
		DE & Deformation & 0.1622 \\
		GR & Branch pipe & 0.0901 \\
		AF & Settled deposits & 0.0811 \\
		\bottomrule
	\end{tabular}
\end{table}

\subsection{Split of Sewer-ML dataset}
Considering the extensive data in the Sewer-ML dataset, it was divided for experimental convenience. The complete Sewer-ML dataset, consisting of 75,618 video crops related to sewer pipelines with expert annotations \cite{Haurum2021}, has a continuous distribution. Images within the same sequence originate from a single video and exhibit a high degree of similarity. In such scenarios, a straightforward image selection strategy, combined with an appropriate dataset size percentage, is essential for facilitating model training and assessing performance improvements. Therefore, a sampling approach was employed, selecting every tenth image to create a smaller dataset.

Given that the training dataset comprises 1,040,129 images and the validation set contains 130,046 images, the authors opted for a one-sixteenth segmentation to balance computational efficiency and dataset sparsity. Further segmentation, such as one-thirty-second, would exacerbate the sparsity issue, especially for defective categories such as IS, FO, PB, OS, and OP, rendering it insufficient for effective model training, as shown in Table 2. Consequently, the chosen one-sixteenth sub-dataset includes 65,008 training images, striking a balance between training time and data adequacy. The validation set is obtained using the same operation, as shown in Table 3.

\begin{table}
	\centering
	\caption{Total number of images for each defect category in the training set for different subsets of the Sewer-ML dataset and the total number of images in the training and validation set.}
	\label{Table2}
	\resizebox{\linewidth}{!}{
		\begin{tabular}{c|ccccccccccccccccc}
			\toprule
			Dataset Size & RB & OB & PF & DE & FS & IS & RO & IN & FO & PH & PB & OS & OP & OK & Normal & Train & Val\\
			\midrule
			1/32 & 1,693 & 6,118 & 264 & 280 & 11,548 & 377 & 842 & 405 & 183 & 663 & 116 & 216 & 61 & 5,668 & 15,572 &-&- \\
			1/16 & 3,299 & 12,849 & 1,189 & 1,045 & 21,488 & 495 & 1,375 & 1,833 & 433 & 1,382 & 295 & 354 & 372 & 11,550 & 30,916 &65,008 & 8,127\\
			1 & 45,821 & 184,379 & 16,254 & 19,084 & 283,983 & 6,271 & 22,637 & 23,782 & 5,010 & 23,685 & 6,746 & 4,625 & 5,325 & 154,624 & 552,820 &1,040,129& 130,046\\
			\bottomrule
	\end{tabular}}
\end{table}

In the sub-dataset, there are only 30,916 normal defect-free images, accounting for 47.56\% of all images, compared to 53.15\% in the entire dataset. So, the evaluation metrics for normal images, ${F1}_{Normal}$, are lower than the results obtained by others when using the sub-dataset in the experimental section. To further compare the distribution of defects in the sub-dataset and the entire dataset, the authors normalize the number of samples for each defect category using the minimum-maximum method. Specifically:
\begin{equation}
	\label{eq30}
	{Value}_{n} = \left( Value \right._{n\_ o} - Min(\bm{Value}))/\left( Max\left( {\bm{Value}} \right) - Min(\bm{Value}) \right)	
\end{equation}
where, \bm{$Value$} is the set of sample numbers of all categories, $Value_{n\_ o}$ and ${Value}_{n}$ represents the sample number of $n$ category before normalization and after normalization, respectively. $Max$ represents to get the maximum value in the set, $Min$ represents to get the minimum value in the set.

Furthermore, as depicted in Fig. 5, the defect distribution in the sub-dataset closely mirrors that of the entire dataset, suggesting that training the model on the sub-dataset is representative of the full dataset. Our focus remains on the training and validation sets since the test set is not publicly accessible.

\begin{figure}[h]
	\centering
	\includegraphics[scale=0.3]{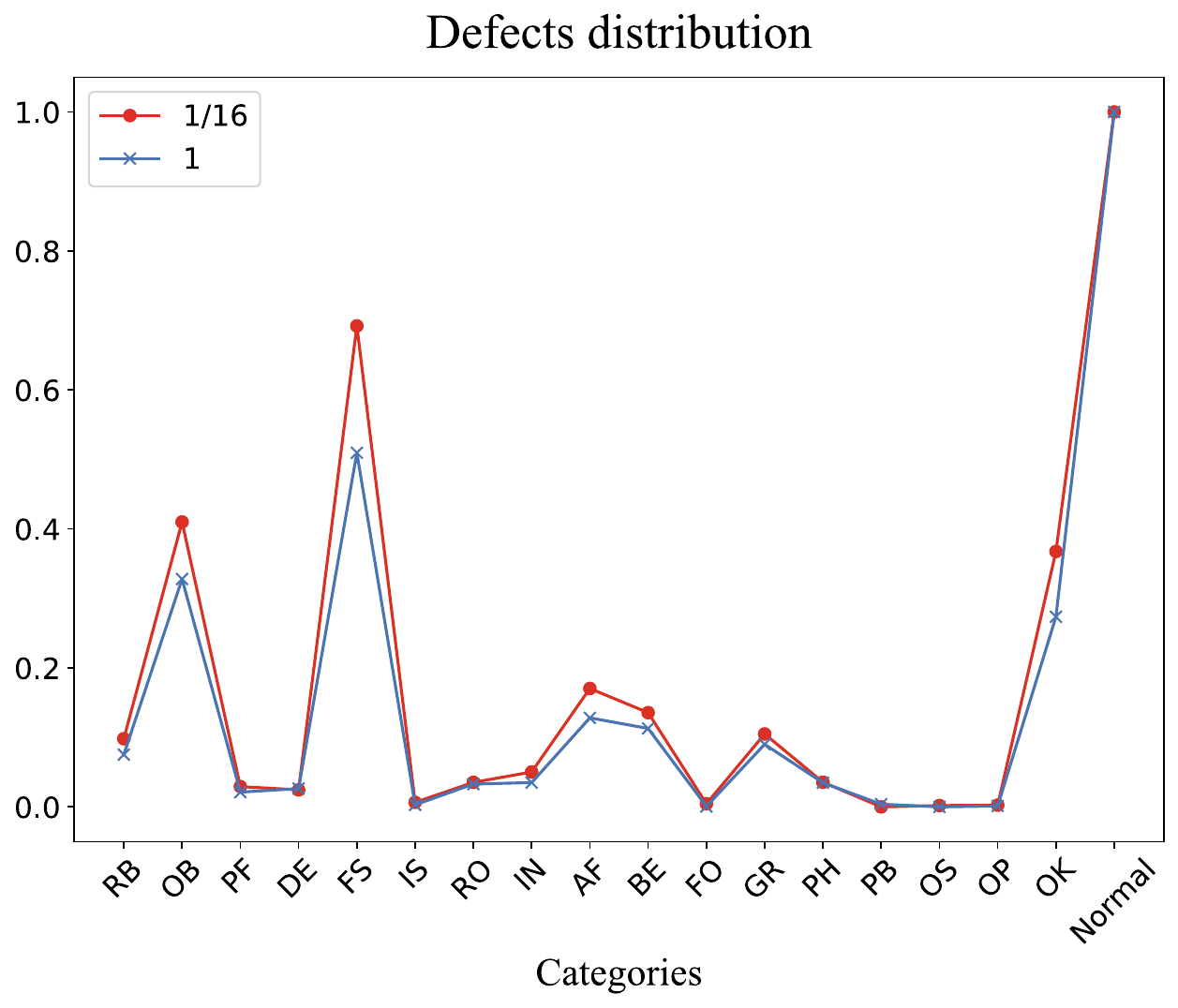}
	\caption{Distribution of the number of defect categories for the different dataset sizes, normalized using max-min.}
	\label{Fig.5}
\end{figure}

\subsection{Selection of bottleneck categories}
\begin{figure}[h]
	\centering
	\includegraphics[scale=0.3]{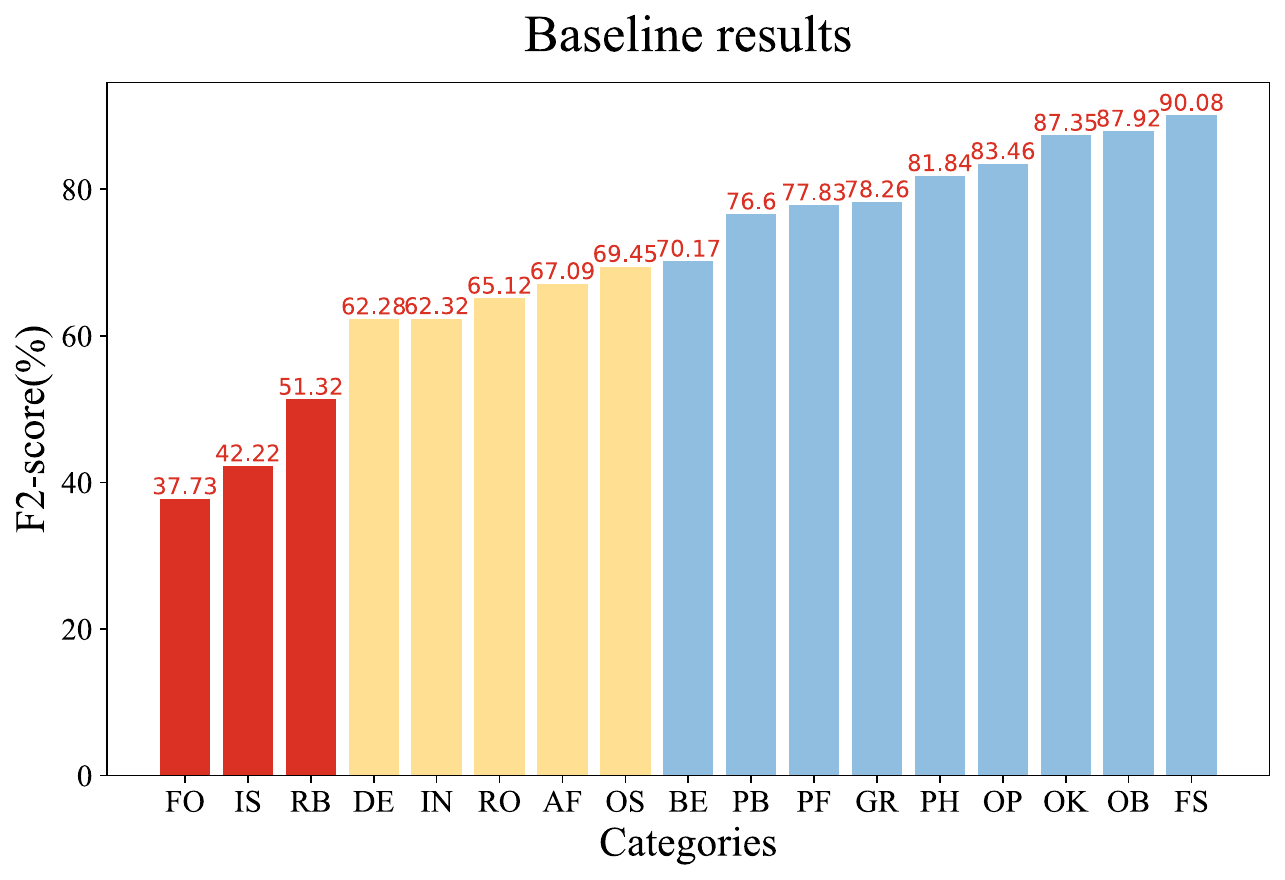}
	\caption{$F2$ scores for different defect categories obtained by training the baseline on the sub-dataset.}
	\label{Fig.6}
\end{figure} 

\begin{figure}[h]
	\centering
	\includegraphics[scale=0.3]{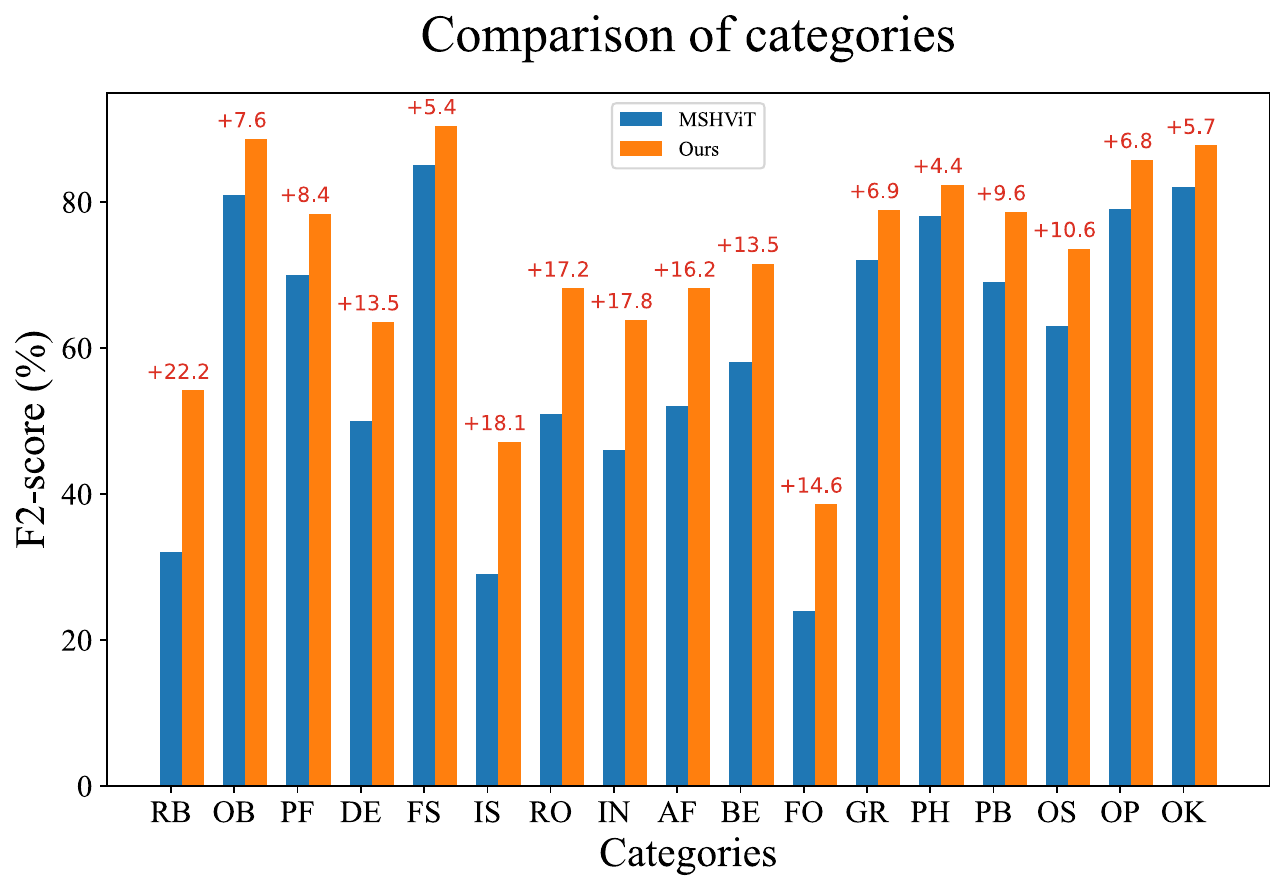}
	\caption{Trained and validated on the entire Sewer-ML dataset to obtain the ${F2}_{CIW}$ for each defect category.}
	\label{Fig.7}
\end{figure}

To effectively address the long-tail problem in sewer pipe defect images, the authors assign higher weight on the bottleneck categories that pose challenges in classification and have a relatively small number of defects. Referring to the best prior results \cite{Haurum2022} on the Sewer-ML dataset, RB, IS, and FO are initially identified as the bottleneck categories, as illustrated in Fig. 7. The validation of this identification is evident when training the baseline model \cite{Liu2021} on the sub-dataset, as shown in Fig. 6. The authors calculate and find that the average value of ${F2}_{CIW}$ for 17 defect categories is 70.06\%. Consequently, the authors focus on the categories with a ${F2}_{CIW}$ value below the average to determine the bottleneck categories.

Aside from the three defect categories with the lowest performance (RB, IS, and FO), there are secondary defect categories (DE, IN, RO, AF, and OS) falling in the 60\%-70\% range. These are considered as potential bottleneck categories. Since our ${F2}_{CIW}$ is computed with the greatest economic impact, the influence of class importance weight on the results must be considered. Among these possible bottleneck categories, the OS has the highest class importance weight, reaching 0.9009. This implies that the detection and classification of the OS defect category significantly impact the performance of the model and indicates a severe economic impact, warranting special attention. Consequently, the authors define RB, IS, FO, and OS as the four bottleneck categories.

\subsection{Comparison with state-of-the-art}

When training and validating on the sub-dataset, our $F2$ metrics reach SOTA, achieving 67.03\%, as shown in Table 4. In addition, considering the potential impact of the incompleteness of the validation set on the final results, the authors also conduct validation on the complete dataset. As the amount of data increases, the ${F2}_{CIW}$ metric decreases but is only 1.0 percentage points worse than the current best result of 63.38\%. This gap is deemed acceptable, given the training dataset reduced by a factor of 16. The lower ${F1}_{Normal}$ is attributed to using the sub-dataset for training, where normal images account for only 47.56\% of the total, compared to other models trained on the entire dataset where normal images account for 53.15\%. When trained on the whole dataset, ${F1}_{Normal}$ improves, outperforming all prior results, indicating that our model is also excellent at recognizing normal images. Notably, the data in Table 4 are all from the original paper, and our model is configured as follows: attention mask, label relevance, and $\alpha = 2$.

It's worth noting that the baseline surpasses the current best result when using only the sub-dataset, showcasing the efficacy of our chosen baseline. Furthermore, our model outperforms the baseline by 3.63 percentage points on the sub-dataset and 1.49 percentage points on the full validation dataset. While the performance of baseline improves with the increase in data on the entire dataset, our model demonstrates greater efficiency, achieving a 1.9 percentage point improvement to 73.15\%, surpassing the prior best result in previous research by 9.77 percentage points.

On the test set, our model continues to excel, surpassing the $F2$ score by 11.41 percentage points. This performance is corroborated by Haurum and Moeslund \cite{Haurum2021}, demonstrating the excellent performance of our model in recognizing and classifying defects in sewer pipe images. It is worth noting that the results on the test set were not provided in the paper by the other authors, so the authors have no way to compare them.

\begin{table}
	\centering
	\caption{Comparison with state-of-the-arts on the validation and test dataset. '*' represents the sub-dataset, otherwise it's on the full validation set. '-' denotes the result is not provided. The bolded number is the best result in each column.}
	\label{Table3}
	\resizebox{\linewidth}{!}{
		\begin{tabular}{cccccc}
			\toprule
			\multirow{2}{*}{Training Dataset} & \multirow{2}{*}{Model} & \multicolumn{2}{c}{Validation} & \multicolumn{2}{c}{Test}\\
			\cmidrule(lr){3-4} \cmidrule(lr){5-6}
			& & F1-score (\%) & F2-score (\%) & F1-score (\%) & F2-score (\%) \\
			\midrule
			\multirow{4}{*}{1/16} & Baseline* \cite{Liu2021} & 89.41 & 63.40 & - & - \\
			& Baseline \cite{Liu2021} & 89.86 & 60.89 & - & - \\
			& Ours* & 89.56 & 67.03 & - & - \\
			& Ours & 89.72 & 62.38 & - & - \\
			\Xhline{0.5pt}
			\multirow{7}{*}{1} & Sewer-ML \cite{Haurum2021} & 91.32 & 55.36 & 90.94 & 55.11 \\
			& CAFEN \cite{Tao2022} & 91.70 & 57.76 & - & - \\
			& CT-GAT \cite{haurum2022multi} & 91.94 & 61.7 & 91.61 & 60.57 \\
			& MSHViT \cite{Haurum2022} & 92.44 & 61.68 & \textbf{92.11} & 60.11 \\
			& SPM \cite{Hu2023} & 91.57 & 63.38 & - & - \\
			& Baseline \cite{Liu2021} & 92.22 & 71.25 & - & -\\
			& Ours & \textbf{92.58} & \textbf{73.15} & 92.10& \textbf{71.98} \\
			\bottomrule
	\end{tabular}}
\end{table}

Furthermore, in comparison to the best model result, MHSViT \cite{Haurum2022}, previously presented for each category ${F2}_{CIW}$, as shown in Fig. 7, our method improves performance across the categories. The most substantial improvement is observed in the bottleneck categories RB and IS, which are the most challenging to recognize, with an improvement of over 20 percentage points on average. This noteworthy and unprecedented improvement underscores that our approach has made significant strides in addressing the issue of category imbalance.

\section{Ablation experiment}
In this section, sub-datasets, including the validation set, are employed for experimental convenience. Firstly, the weights are determined by ablation experiments based on the selection of the bottleneck categories in section 4.3. Subsequently, the correctness of the bottleneck category selection is experimentally demonstrated. Next, the authors present the different backbone options based on baseline and discuss the specific performance of each component of our improved structure. Finally, experimental details about dynamic weight values are presented.

\subsection{Static value weight}
\begin{figure}[h]
	\centering
	\includegraphics[scale=0.3]{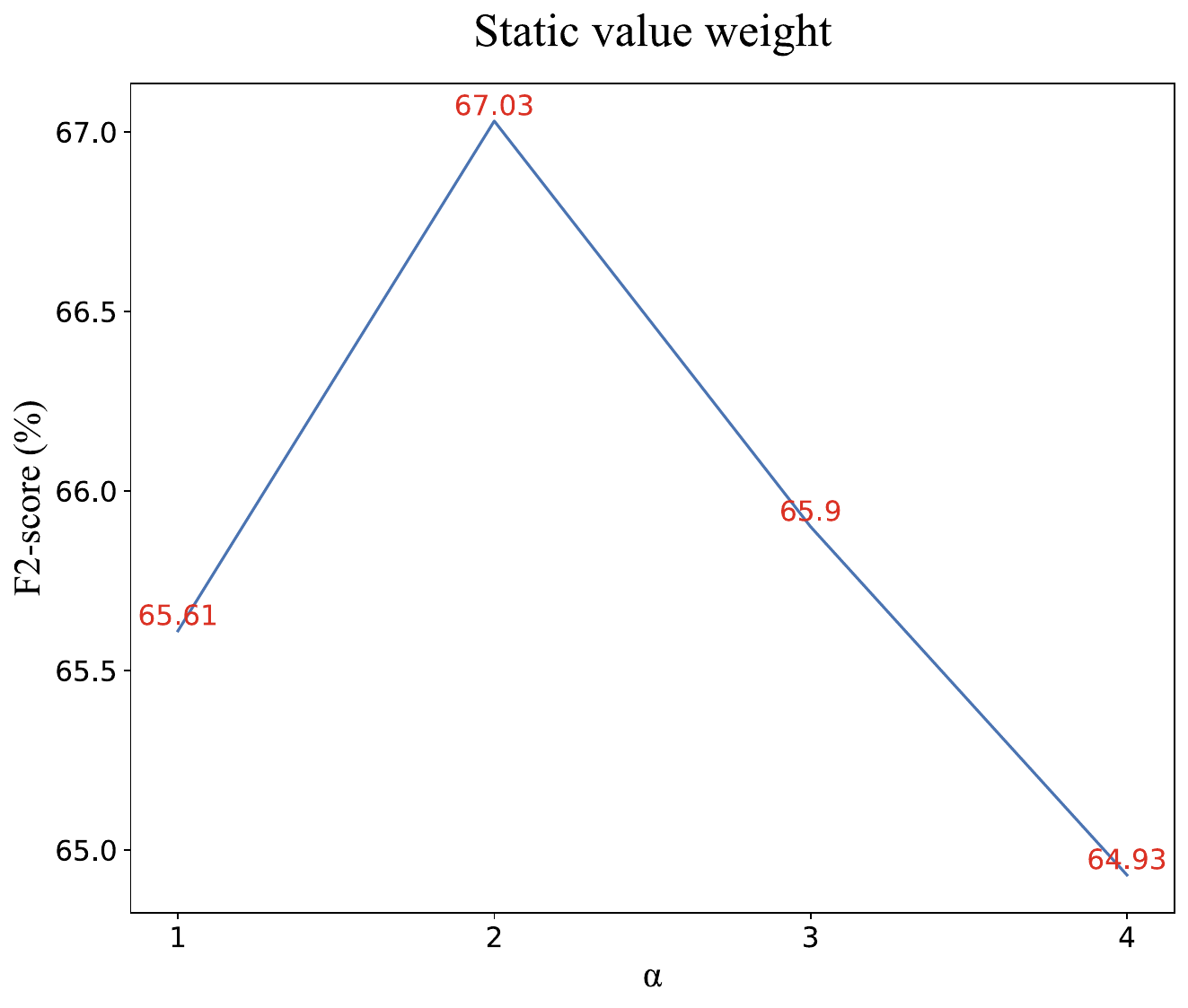}
	\caption{The varied ${F2}_{CIW}$ results for different values of $\alpha$ are trained and validated on the sub-dataset.}
	\label{Fig.8}
\end{figure} 

Based on our theoretical analysis in section 4.3, the authors select RB, IS, FO, and OS as our bottleneck categories. By enumerating several values, it is observed that the weight value $\alpha = 2$, ${F2}_{CIW}$, produces the best results. This demonstrates that increasing the weight value effectively allows the model to learn the bottleneck categories in a targeted manner, thereby enhancing the overall performance of the model, as shown in Fig. 8. When $\alpha = 1$, without increasing the weights, the model is not allowed to target learning. It is noteworthy that when $\alpha$ exceeds 2, ${F2}_{CIW}$ begins to gradually decrease, suggesting that imposing too large weight value will cause the model to focus too much on the bottleneck categories and neglect the learning of other defective categories, which is undesirable. Therefore, the authors decided to set the weight value $\alpha$ to 2, a static value, as the final model parameter, allowing us to target the bottleneck categories while still retaining sufficient learning for other defective categories to achieve the best performance.

\subsection{Verification of bottleneck categories}
\begin{table}
	\centering
	\caption{Experiments on various sets of bottleneck categories are trained and validated on the sub-dataset.}
	\label{Table4}
	\begin{tabular}{ccc}
		\toprule
		Bottleneck Categories & F1-score (\%) & F2-score (\%)\\
		\midrule
		RB+IS+FO & 89.73 & 66.32 \\
		RB+IS+FO+DE & 89.63 & 66.11 \\
		RB+IS+FO+RO & \textbf{89.79} & 66.73 \\
		RB+IS+FO+IN & 89.57& 65.1 \\
		RB+IS+FO+AF & 89.62 & 65.69 \\
		RB+IS+FO+OS & 89.56 & \textbf{67.03} \\
		\bottomrule
	\end{tabular}
\end{table}
In order to verify our theory about the bottleneck categories, the authors conduct weighted experiments on different combinations of categories. Initially, the three categories are validated which are the most difficult to categorize in prior study results: RB, IS, and FO. However, the results were not satisfactory, as shown in Table 4. Subsequently, according to our theory in section 4.3, the authors add DE, RO, IN, AF and OS to the set of bottleneck categories respectively. The results support our theory: since OS has the largest proportion in the CIW, it has the greatest impact on the results when calculating the ${F2}_{CIW}$, leading to improved model performance. These experimental results demonstrate that our selection of bottleneck categories and weighting strategy is acceptable and proper, capable of achieving considerable performance improvement on sewer pipe defect data. By focusing on the bottleneck categories for enhanced learning, our model effectively addresses the category imbalance problem of the dataset and achieves excellent classification performance.

\subsection{Backbone}
The authors select the recent better performing multi-label classification models \cite{Lanchantin2021}, \cite{Liu2021}, \cite{Ridnik2021tresnet}, and find that Q2L-resnet101 achieved optimal performance, even when the other model has larger input size, as shown in Table 5. Therefore, Q2L is selected as our baseline. To explore the effect of different backbones on the model performance, the authors adopt \cite{Ridnik2021tresnet}, \cite{Ryali2023}, \cite{Wu2021}, \cite{liu2021swin} as the backbone and follow the configuration in the corresponding paper. It is worth noting that due to the different backbone settings, the model also has different input image sizes. Larger image input can provide more image information, but even with the model using Tresnet-L as the backbone, the performance with an input image size of 448 $\times$ 448 was not as good as expected. On the other hand, for smaller sizes of the input image, the performance difference with the model employing resnet101 as the backbone is significant, and the size of the input image is not a key factor in bridging this performance gap. In addition, from the comparison of the number of parameters, the model with resnet101 has the lowest number. Considering the above factors, the authors  finally choose the Q2L model with resnet101 and with an input size of 448 $\times$ 448 as our baseline, as it achieved better performance results while keeping the number of parameters relatively low.

\begin{table}
	\centering
	\caption{Different model and backbone are trained and validated on the sub-dataset.}
	\label{Table5}
	\resizebox{\linewidth}{!}{
		\begin{tabular}{ccccccc}
			\toprule
			& Model & Input Size & Params (M) & mAP (\%) & F1-score (\%) & F2-score (\%) \\
			\midrule
			\multirow{3}{*}{Different Model} & TResNet-L \cite{Ridnik2021tresnet}  &  448 $\times$ 448 & 53 & 44.53 & 90.33 & 51.7 \\
			& Q2L-Resnet101 \cite{He2016}  & 448 $\times$ 448 & 143 & 60.07 & 89.41 & \textbf{63.4} \\
			& C-Tran \cite{Lanchantin2021}  & 576 $\times$ 576 & 68 & 59.77 & 89.55 & 52.4\\
			\Xhline{0.5pt}
			\multirow{4}{*}{Different Backbone} & Q2L-Hiera-L \cite{Ryali2023}  & 224 $\times$ 224 & 268 & 44.24 & 84.81 & 49.62 \\
			& Q2L-CvT-w24 \cite{Wu2021}  & 224 $\times$ 224 & 326 & 46.49 & 86.7 & 51.47 \\
			& Q2L-Swin-L \cite{liu2021swin}  & 384 $\times$ 384 & 270 & 48.49 & 86.64 & 52.26 \\
			& Q2L-TResnet-L \cite{Ridnik2021tresnet}  & 448 $\times$ 448 & 173 & 43.84 & 83.10 & 50.06 \\
			\bottomrule
	\end{tabular}}
\end{table}

\subsection{Effect of each module}
The authors verify the impact of each part of the model on the performance by adding modules or structures independently step by step. In Table 6, the original baseline achieved 63.4\% performance, which is already the current optimal result. This indicates that the baseline the authors chose is very suitable for the task of sewage pipe defect identification and classification. 

With the addition of the attention mask, the performance is improved by 2.36 percentage points, which is the most significant improvement in our improved module. This result proves that the attention mask generated through the class activation map is very effective in mining and utilizing the defective local spatial discriminative information. And, incorporating self-attention computation into the decoder improves the performance by 1.54 percentage points. Due to the coexistence of multiple classes of defects in the image, the model can better capture the co-occurrence relationship between different labels for more accurate classification through self-attention computation. Furthermore, the enhanced learning of the bottleneck categories results in a 1.82 percentage point improvement in the overall performance of the model, further addressing the category imbalance of the data. Finally, integrating all three aspects of our improvement together, the overall performance reaches 67.03\%, achieving a 3.61 percentage point improvement compared to the baseline. This fully proves that our improvement is effective. Previous research has found that specific defects in sewer pipe images are especially difficult to recognize, making the images challenging to apply in real-world applications. Our enhancement has significantly increased performance and advanced the modeling application process.

\begin{table}
	\centering
	\caption{The effects of different module structures on the F1-scores and F2-scores. The best results are highlighted in bold.}
	\label{Table6}
	\resizebox{\linewidth}{!}{
	\begin{tabular}{ccc}
		\toprule
		Model & F1-score (\%) & F2-score (\%) \\
		\midrule
		Baseline & 89.41 & 63.4 \\
		+ attention mask & 89.79 & 65.76 \\
		+ self-attention & 89.47 & 64.94 \\
		+ weighted bottleneck category & 89.27 & 65.22 \\
		+ attention mask + self-attention + weighted bottleneck category  & 89.56 & \textbf{67.03} \\
		\bottomrule
	\end{tabular}
}
\end{table}

\subsection{Dynamic value weight}

\begin{figure}[h]
	\centering
	\includegraphics[scale=0.3]{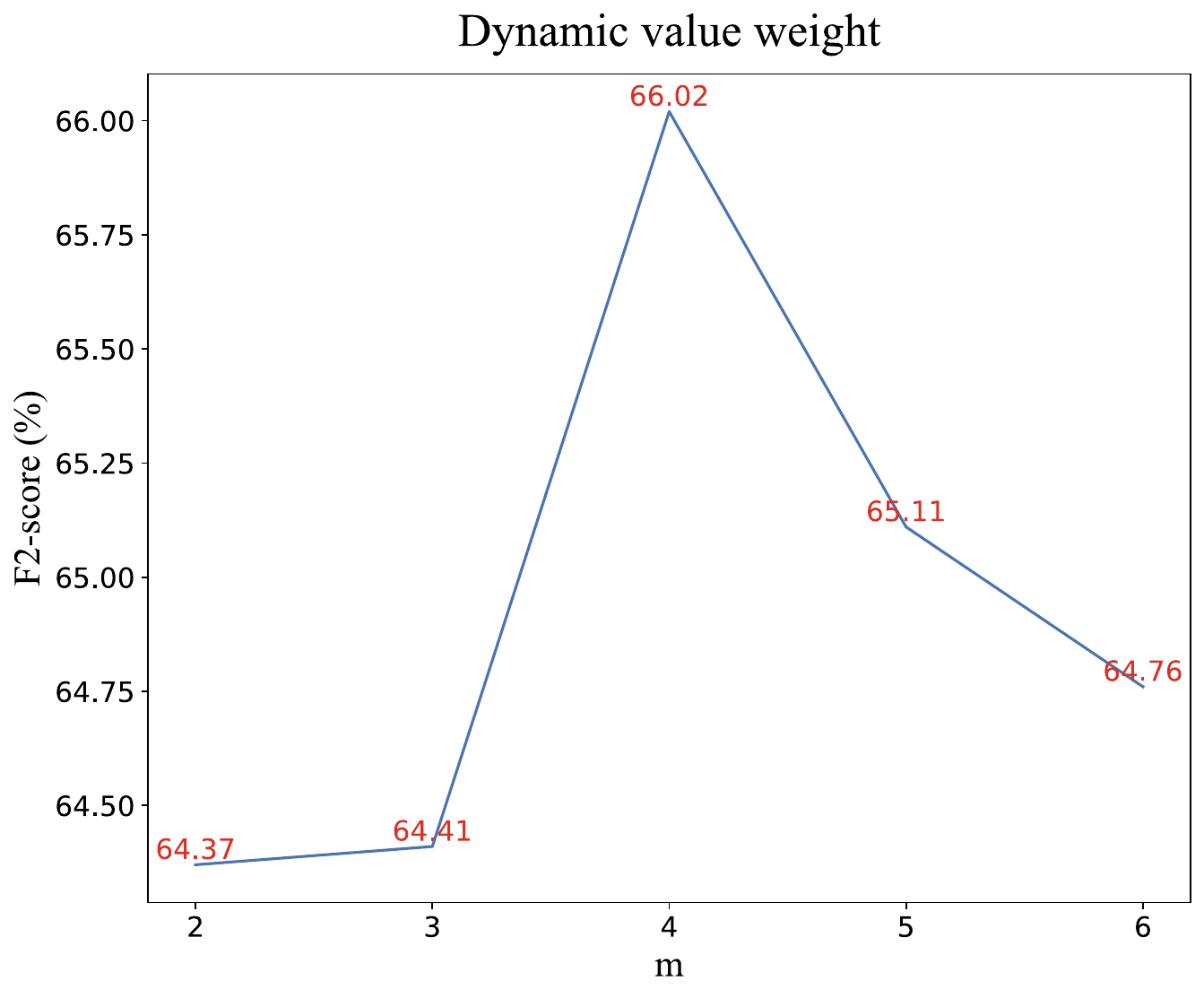}
	\caption{Training and validation on the sub-dataset in the manner described in 3.6 for dynamical weight values. $m$ represents the top $m$ worst performance categories selected by the model as the bottleneck categories for enhancement learning.}
	\label{Fig.9}
\end{figure} 
In order to explore the impact of different values of $\alpha$ on the model performance, the authors designed a dynamic way of determining $\alpha$. Following section 3.6, the authors experimented with different values of $m$ to allow the model to dynamically select the set of bottleneck categories that need to be augmented for learning during the training process. As shown in Fig. 9, our model achieves the best performance when $m = 4$, demonstrating that using four classes as the bottleneck categories is the best combination result. This also verifies that our analysis of the bottleneck categories and the number of selections in section 4.3 is correct. However, the outcomes of weighing using dynamically generated weight values are inferior to those produced using static values. The main reason for this might be that the weight value $\alpha$ is calculated based on the $mAP$ of each training iteration, and the weight value keeps changing. At the beginning of the training, it is possible that the bottleneck categories are not weighted, but rather other categories with high confidence originally, which may cause the model to ignore the bottleneck categories even more. As the number of iterations increases, the model gradually focuses on the bottleneck categories. When training converges, it is likely that the model has not sufficiently learned the defective features of the bottleneck categories, leading to a lower overall accuracy. Therefore, the authors finally decided to use static values instead of dynamic values to ensure that the model focuses on the bottleneck categories throughout the training process and maintains consistent weight assignments in all situations, which can improve the overall performance.

\subsection{Different input size}

\begin{figure}[h]
	\centering
	\includegraphics[scale=0.3]{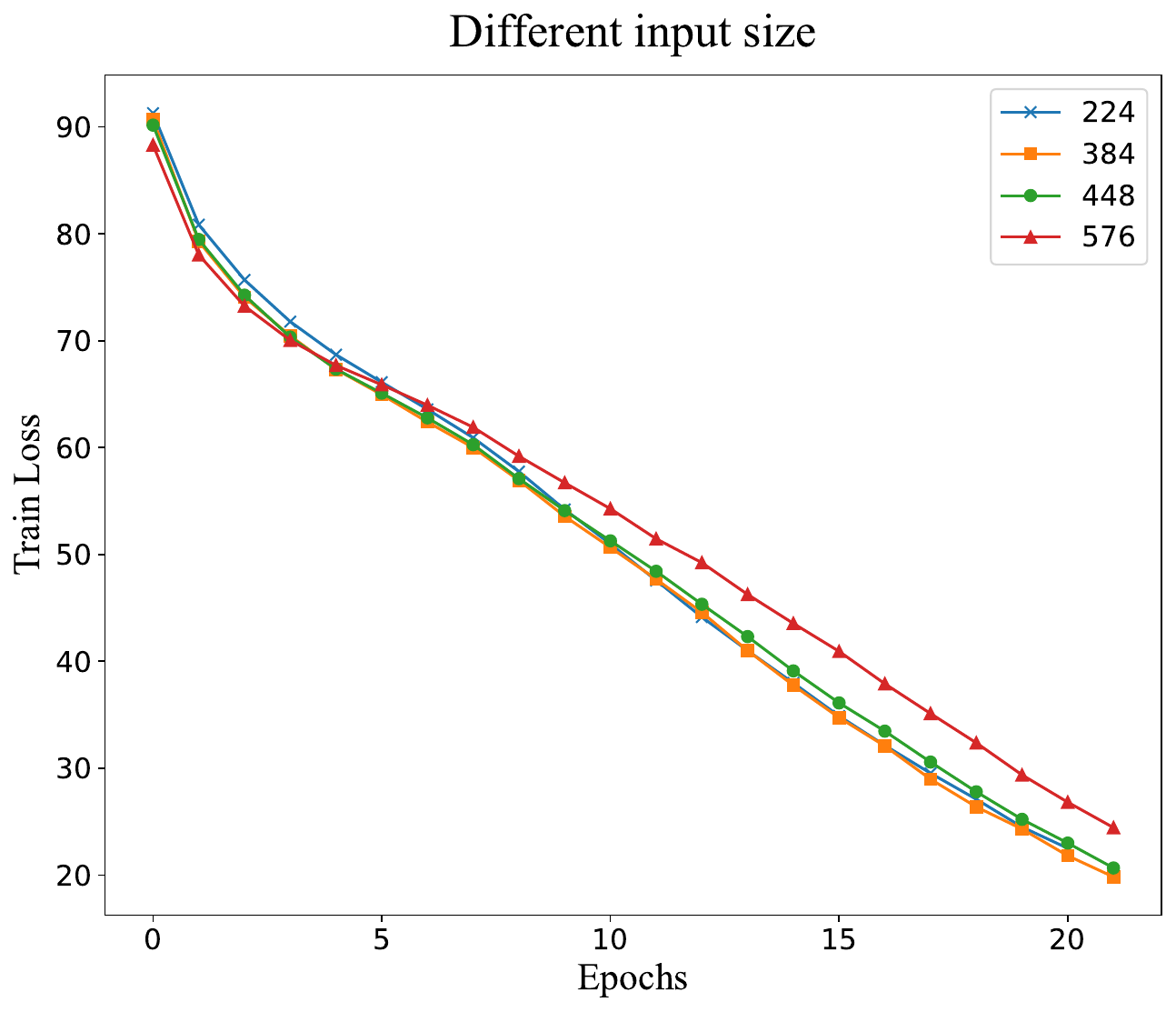}
	\caption{ As the training dataset undergoes transformations at different resolutions, variations in the training loss across iterations are observed.}  
	\label{Fig.10}
\end{figure} 

\begin{figure}[h]
	\centering
	\includegraphics[scale=0.3]{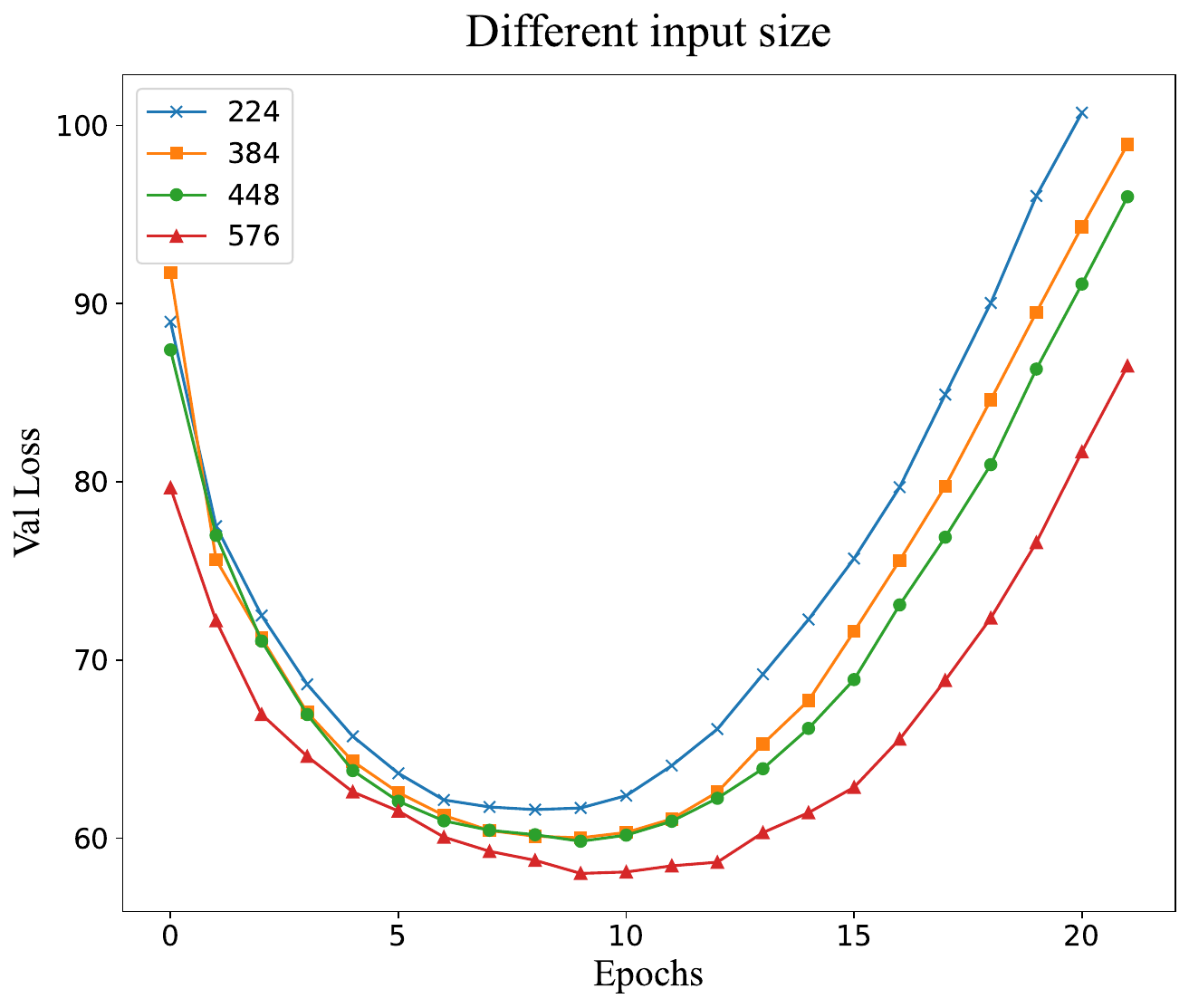}
	\caption{ As the validation dataset undergoes transformations at different resolutions, variations in the training loss across iterations are observed.}  
	\label{Fig.11}
\end{figure}

\begin{figure}[h]
	\centering
	\includegraphics[scale=0.3]{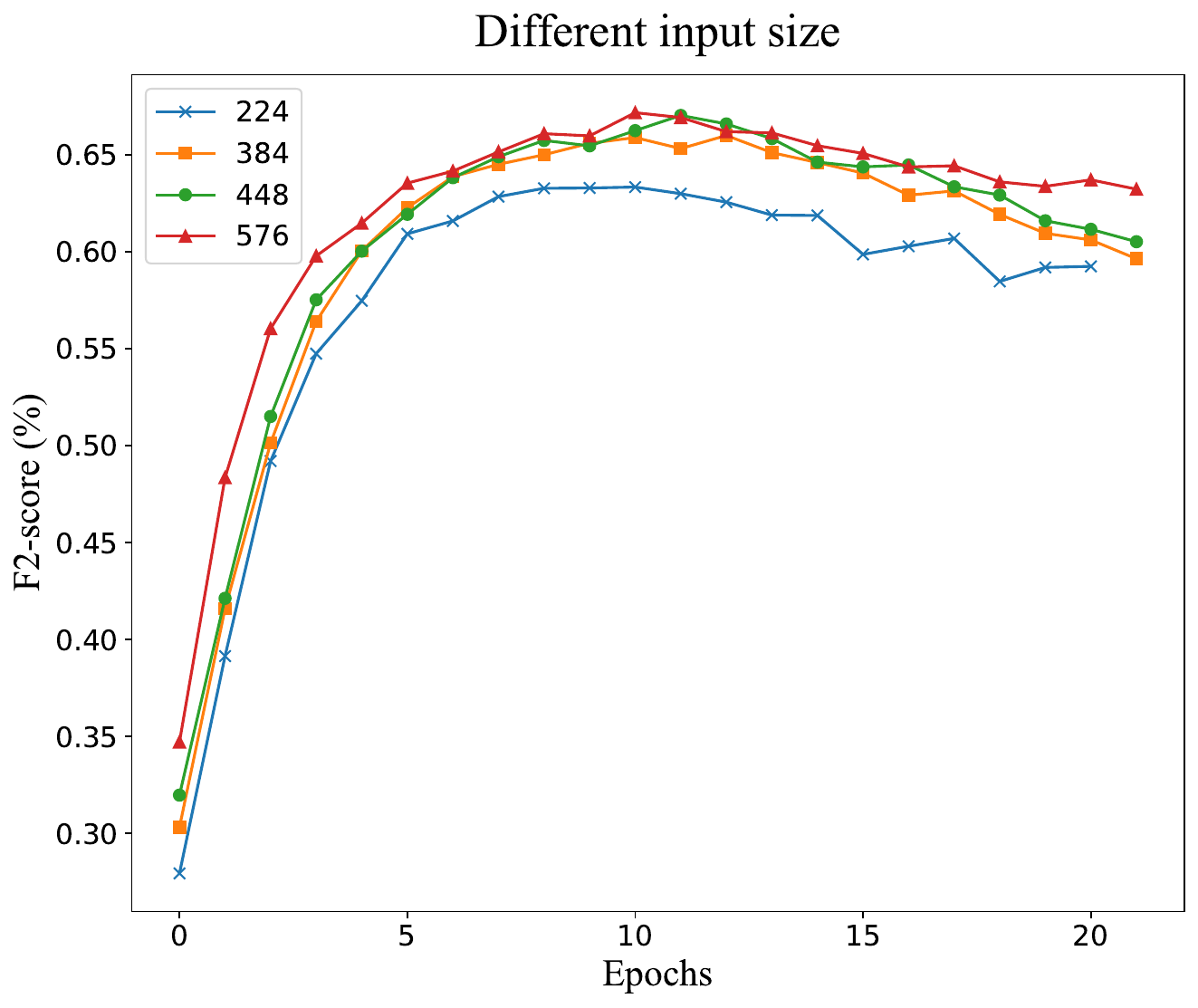}
	\caption{  The F2-score of the validation dataset at various resolutions changes with the training iterations.}  
	\label{Fig.12}
\end{figure}

\begin{table}
	\centering
	\caption{Results on sub-validation datasets with different input sizes on our model.}
	\label{Table7}
	\begin{tabular}{ccccc}
		\toprule
		Input Size & Params (M) & FLOPs (G) & F1-score (\%) & F2-score (\%) \\
		\midrule
		224 $\times$ 224 & 143 & 19.58 & 89.25 & 63.34\\
		384 $\times$ 384 & 143 & 28.89 & 89.37 & 66.00\\
		448 $\times$ 448 & 143 & 38.91 & 89.56 & 67.03\\
		576 $\times$ 576 & 143 & 63.59 & 89.68 & 67.17\\
		\bottomrule
	\end{tabular}
\end{table}

To explore the relationship between different resolutions and accuracy, experiments were conducted using various input scales to train the model. Loss comparisons and accuracy changes for training at different resolutions were provided. For experimental convenience, experiments were conducted on sub-datasets. Overfitting in the model was observed based on the training and validation losses, in Fig. 10 and Fig. 11. To address this issue, the same strategy as Q2L\cite{Liu2021} was employed, which involved early stopping. Given the large number of images in the dataset, the model tends to learn more information per epoch, resulting in rapid convergence. Early stopping is typically triggered after training for approximately 20 epochs.
Images of different resolutions contained varying amounts of information. The larger the resolution, the more information the model could potentially learn. As shown in Fig. 10, the training losses for four different scales were initially sorted according to scale size. However, it was important to note that a larger scale also meant the model was more likely to capture noise, leading to increased computational complexity. As the training epoch progressed, it became evident that the loss values tended to increase with larger scales, as depicted in Figure 10. This phenomenon could be attributed to increased noise interference, necessitating longer training iterations.
However, despite encountering noise at larger scales, significant performance improvements were observed in Fig. 11 and Fig. 12. These improvements manifested as lower losses on the validation set and higher ${F2}_{CIW}$ indices.
Additionally, considering that the training loss for the 448 $\times$ 448 scale was lower than that for the 576 $\times$ 576 scale, indicating relative stability, and the latter incurred nearly double the computational cost, the modest 0.14\% improvement in accuracy did not justify the increased computational burden. Therefore, the model finally adopted an input size of 448 $\times$ 448.

\section{Visualization and defect localization}

In this section, the visualization analysis is first performed for the bottleneck categories. Subsequently, eight defective categories are selected for visualization to further underscore the effectiveness and the interpretability of our approach in mining local spatial features.

\subsection{Visualization of bottleneck categories}

\begin{figure}[h]
	\centering
	\includegraphics[width=\textwidth]{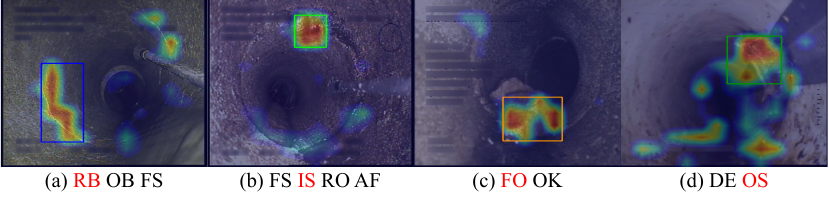}
	\caption{ Visualization of specified defect categories, with specified items in red. RB:Cracks, breaks, and collapses; OB:Surface damage; FS:Displaced joint; IS:Intruding sealing material; RO:Roots; AF:Settled deposits; FO:Obstacle; OK:Connection with construction changes; DE:Deformation; OS:Lateral reinstatement cuts.}
	\label{Fig.13}
\end{figure} 

Our model offers sewer pipe repairers with a visual representation of a rough heatmap of image defects. The matrix within this heatmap is adjusted by applying a threshold to the weight values in the class activation map, selecting only values surpassing the threshold, and setting the rest to zero. Then, non-zero values are binarized to 1, and object contours are extracted from the map using OpenCV library methods. The resulting matrix is further filtered based on area judgments.

The heatmap results for displaying the four bottleneck categories are presented in Fig. 13. The model effectively identifies distinct defect categories even when they co-exist with other defects and are located at uncertain spatial positions in the image. However, some issues were identified through the rough defect localization map. Since the images are selected frame by frame from the video, the limited illumination area in the video can lead to the deepest part of the pipe not being well-lit. This may cause the model to mistakenly infer the presence of defects, especially for OS defect, located on the inner wall of the pipe closer to the darker area. This scenario makes it challenging for the model to focus its attention, leading to distractions and false positives, as observed in Fig. 13(d).

Although our model can effectively and roughly localize the area showing the defects, challenges persist due to the limitation of the illuminated area by captured light. Beyond our work, addressing this peculiarity may further enhance the recognition of defective images in sewer pipes, and this will be a focus of our future work.

\subsection{Defects Localization}
Our model roughly localizes the areas of defect categories by means of heat maps. The authors give some examples of common defect combinations to show the ability of our model to roughly localize defects regions, as shown in Fig. 14.

\begin{figure}[h]
	\centering
	\includegraphics[width=\textwidth]{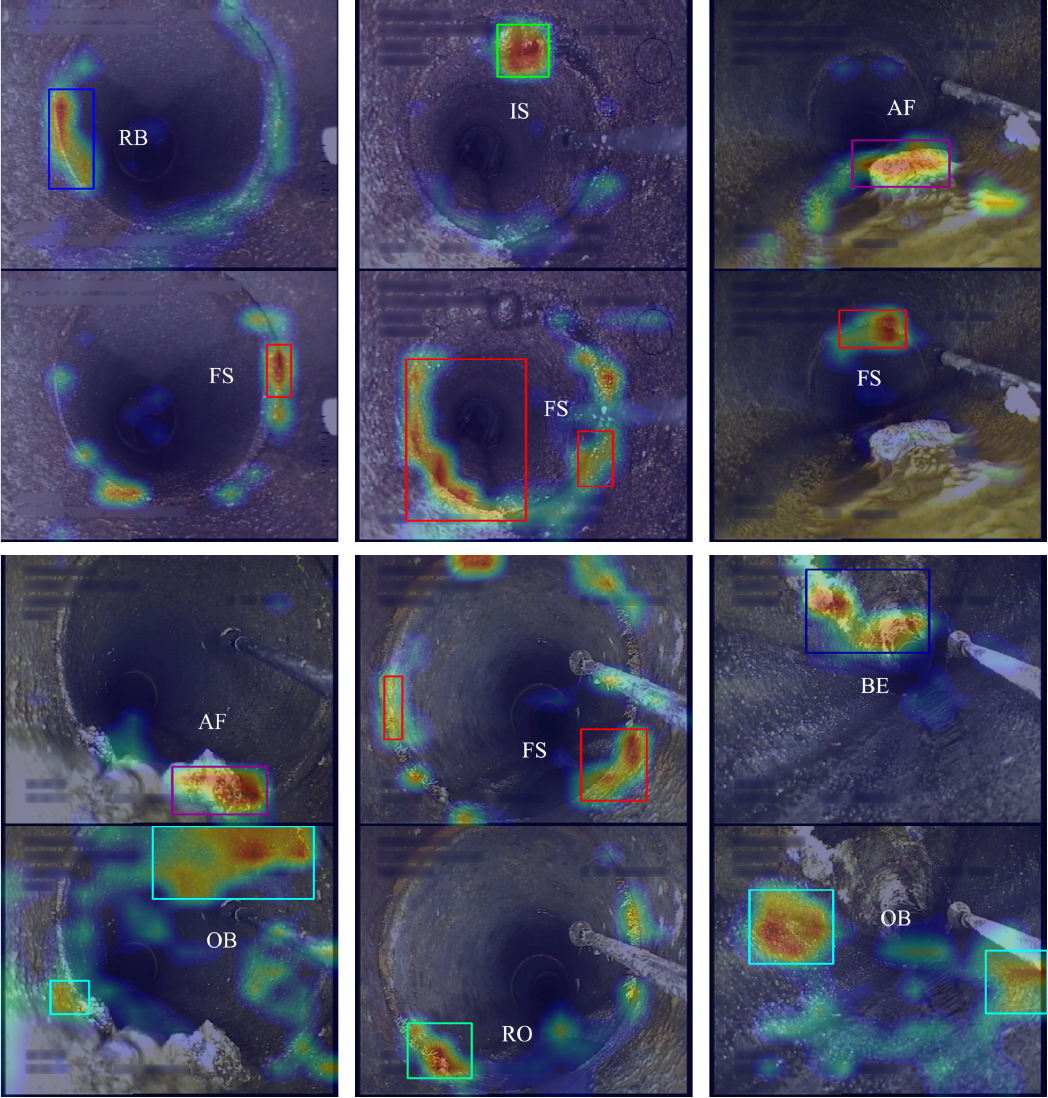}
	\caption{ Rough defect localization heatmaps for different defect categories in the same image. RB:Cracks, breaks, and collapses; IS:Intruding sealing material; AF:Settled deposits; FS:Displaced joint; BE:Attached deposits; OB:Surface damage; RO:Roots;}  
	\label{Fig.14}
\end{figure}

\section{Conclusion}
In order to deeply mine and utilize the local spatial feature information of defects, the authors adopt the class activation map to generate the local spatial discriminative information of defects as the attention mask and participate in the attention computation to achieve feature enhancement. Also, for the coexistence of multi-class defects and serious class imbalance problems in the sewer pipe defective images, the authors adopt the self-attention mechanism of label embedding to learn the relationship between labels and enhance the label relevance. In addition, the authors use a weight updating strategy based on an asymmetric loss function for bottleneck categories enhancement learning to further address the category imbalance problem in the data. Since the dynamic value approach fails to show as much superiority as the static value approach, the authors ultimately choose to use the static-value weight update strategy. Our model achieves results approximating SOTA on just one-sixteenth of the Sewer-ML dataset, and outperforms the current best method $F2$ performance by 11.87\% on the full dataset. In addition, our model can provide rough heatmaps for defect localization to workers repairing pipes, and the authors hope that our work will advance the field of defect recognition in sewer pipes. Our future work may further investigate this area from the perspective of the wrong region of interest due to insufficient light deep in the pipe.

\section{Data Availability Statement}
Some or all data, models, or code that support the findings of this study are available from the corresponding author upon reasonable request.

\section*{Acknowledgement}
This work was supported in part by NSF of China under Grant No. 61903164 and in part by NSF of Jiangsu Province in China under Grants BK20191427.

\bibliography{reference}


\end{document}